\documentclass{WileyMSP-template}

\usepackage{amsmath}
\usepackage{amsthm}
\usepackage{mathtools}

\makeatletter
\renewcommand{\fnum@figure}{\textbf{Figure \thefigure}}
\renewcommand{\fnum@table}{\textbf{Table \thetable}}
\makeatother

\usepackage{newtxtext,newtxmath}

\usepackage{xr}
\usepackage{commath,graphicx,afterpage,caption,subfigure,amsmath,algorithmic,mathtools}
\usepackage{amsmath,cite,multirow,multicol,booktabs,threeparttable,extpfeil}
\usepackage{amsthm,hyperref}
\usepackage[mathlines]{lineno}
\usepackage{etoolbox}
\usepackage[most]{tcolorbox}
\usepackage{tikz}
\definecolor{orcidgreen}{HTML}{A6CE39}
\newcommand{\orcidID}[1]{\,\href{https://orcid.org/#1}{\tikz[baseline=-0.45ex]{\node[circle,fill=orcidgreen,text=white,minimum size=1.55ex,inner sep=0pt,font=\sffamily\bfseries\fontsize{5}{5}\selectfont]{iD};}}}

\newtheorem{theorem}{Theorem}
\newtheorem{assumption}{Assumption}
\newtheorem{lemma}{Lemma}

\theoremstyle{definition}

\DeclareMathOperator*{\argmax}{arg\,max}

\usepackage{ifpdf}
\ifpdf\else
  \AtBeginDocument{\special{! << /AutoRotatePages /None >> setdistillerparams}}%
\fi

\begin{document}

\pagestyle{fancy}

\title{StackingNet: Collective Inference across Independent AI \\ Foundation Models}

\maketitle

\author{Siyang Li$^{1\dagger}$\orcidID{0009-0009-4874-3501}}
\author{Chenhao Liu$^{1\dagger}$\orcidID{0009-0007-4025-4389}}
\author{Dongrui Wu$^{1\ast}$\orcidID{0000-0002-7153-9703}}
\author{Zhigang Zeng$^{1\ast}$\orcidID{0000-0003-4587-3588}}
\author{Lieyun Ding$^{2\ast}$\orcidID{0000-0002-9873-3776}}

\dedication{$^\dagger$These authors contributed equally to this work.}

\begin{affiliations}
$^{1}$School of Artificial Intelligence and Automation, Huazhong University of Science and Technology, Wuhan 430074, China.\\
$^{2}$School of Civil and Hydraulic Engineering, Huazhong University of Science and Technology, Wuhan 430074, China.\\
$^\ast$Corresponding authors. Email: drwu@hust.edu.cn, zgzeng@hust.edu.cn, dly@hust.edu.cn
\end{affiliations}


\begin{abstract}
Artificial intelligence built on large foundation models has transformed language understanding, computer vision, and reasoning, yet these systems remain isolated and cannot readily share their capabilities. Coordinating the complementary strengths of independently developed, black-box foundation models is essential for trustworthy intelligent systems, yet no established method exists. Here we show that such coordination can be achieved through a meta-ensemble framework termed StackingNet, which aggregates the output predictions of independent models at inference. StackingNet improves accuracy, reduces individual-model error and group-wise disparities, ranks model reliability, and identifies or prunes models that degrade performance, all without access to internal parameters or training data. Across language comprehension, visual attribute estimation, and academic paper rating, it consistently outperforms individual models and classic ensembles, with gains that persist when the base models are uniformly strong. These gains stem from variance reduction and consensus alignment among independent models rather than from any emergent group cognition, and they widen as the model pool grows more diverse. By turning model diversity from a source of inconsistency into a resource for cooperation, StackingNet offers a practical path toward coordinated artificial intelligence, where progress emerges not only from larger single models but from principled cooperation among many specialized ones.
\end{abstract}

\section*{Introduction}

Foundation models have transformed artificial intelligence (AI). Large language models (LLMs)~\cite{Zhao2023} and vision-language models (VLMs)~\cite{Zhang2024}, trained on vast and diverse datasets, now exhibit strong generalization across a wide range of tasks. These billion-parameter models now form the basis of modern AI systems, enabling advances in language understanding~\cite{Hendrycks2021}, visual reasoning~\cite{Antol2015}, and many more related domains. Their applications have further expanded to automatic code generation~\cite{Chen2021LLMcode, Liu2023}, academic writing and assistance~\cite{Milano2023}, large-scale data annotation~\cite{Tan2024}, and even research paper review~\cite{Zhuang2025, Liang2024LLMpeerreview}, reflecting their growing role in everyday computational and creative work.

Most foundation models are developed by large organizations with access to proprietary data and extensive computational resources. They are typically deployed as isolated black boxes, offering little transparency regarding their training data, architectures, or optimization objectives. Their internal mechanisms remain opaque, and their biases and failure modes are poorly understood. These limitations raise concerns about explainability~\cite{Zhao2024}, trustworthiness~\cite{Liang2022}, and fairness~\cite{Jobin2019, Gallegos2024, Liang2021LLMbias}. Efforts in model alignment, including instruction tuning~\cite{Ouyang2022} and value alignment with human norms and legal principles~\cite{Ji2023, Jobin2019}, have improved the ethics of AI but not their coordination.

This absence of coordination contrasts with collective behavior in biological and social systems. In animal groups such as fish schools, bird flocks and wolf packs, reliable group behavior arises from local interactions among individuals~\cite{Duan2023}. Human decision-making shows similar patterns through voting, collaboration, and consensus building, where individual judgments are combined to improve accuracy and fairness. These examples suggest that reliable outcomes should and must emerge from a group of independent individuals under appropriate aggregation mechanisms. Modern AI ecosystems, with models now reaching trillions of parameters~\cite{Ren2023}, resemble such intelligent systems only in scale, but not in coordination.

Ensemble learning offers a computational foundation for such aggregation. However, traditional ensemble methods are built during model training~\cite{Cao2020} and cannot be directly applied to pre-trained models. To bridge this gap, we draw on combination methods originally developed for crowdsourcing~\cite{Li2016}. Early frameworks such as the Dawid-Skene model~\cite{Dawid1979} and stacking~\cite{Wolpert1992} showed that model predictions alone can be combined effectively, even without access to training data or internal parameters.

\begin{figure}[htpb]\centering
\includegraphics[width=\textwidth]{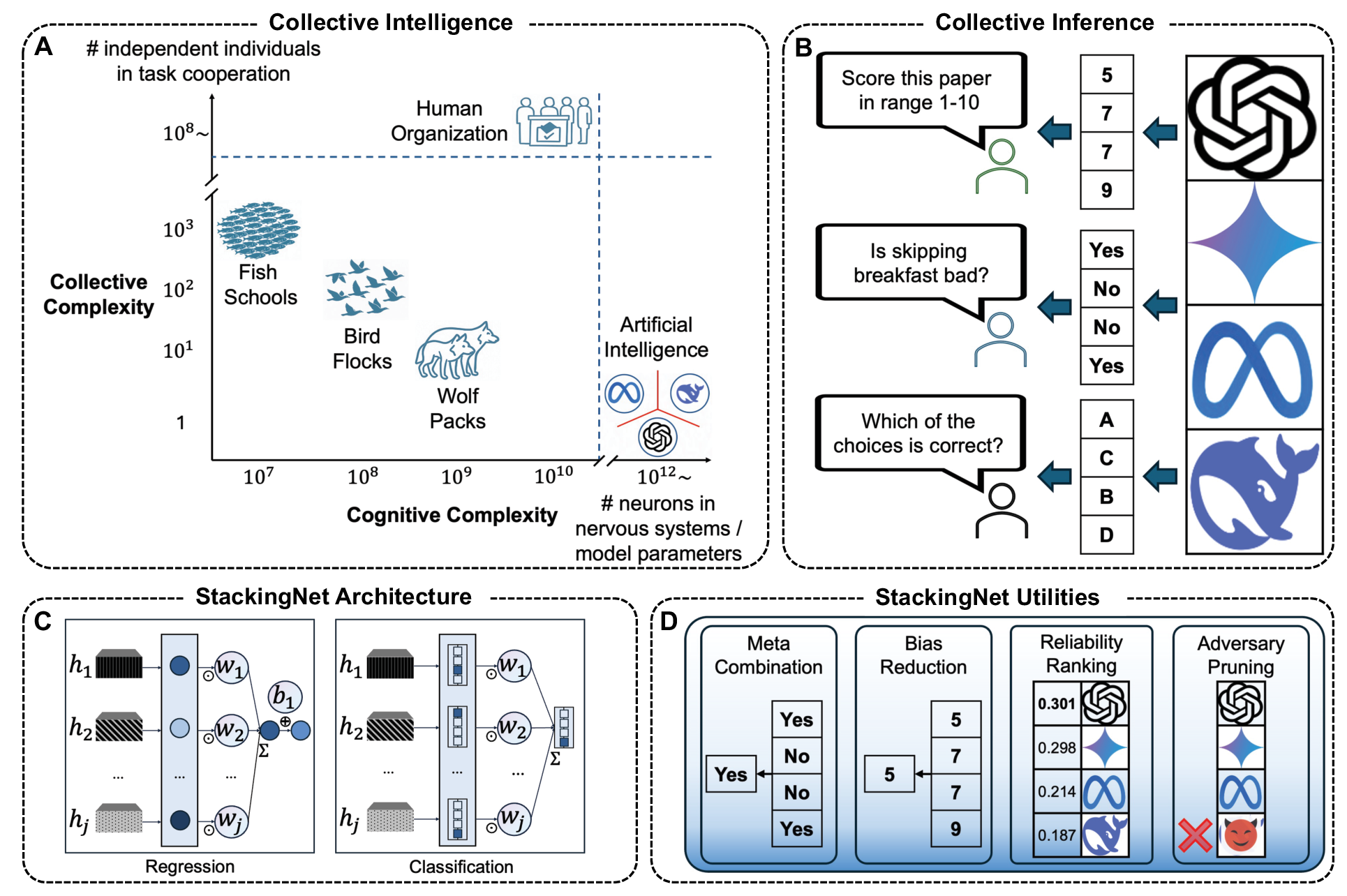}
\caption{{\bfseries Collective intelligence and collective inference across independent intelligent systems.}
{\bfseries a}, Illustrative analogy of collective intelligence across biological and artificial systems, positioned by cognitive complexity (the number of neurons or model parameters, horizontal axis) and collective complexity (the number of independent individuals cooperating on a task, vertical axis).
{\bfseries b}, Collective inference, the post-hoc aggregation of the output predictions of multiple independent foundation models across diverse task types, performed here by StackingNet.
{\bfseries c}, StackingNet architecture and its learnable parameters, combining each black-box base model through a trainable meta-learner for regression or classification.
{\bfseries d}, Functional utilities of StackingNet: combining model outputs, reducing individual-model error and group-wise disparities, estimating reliability with or without supervision, and filtering unreliable or adversarial models.
Reliability scores are calculated based on publicly available benchmark LMarena (\url{https://lmarena.ai/leaderboard}), accessed on Oct. 5, 2025. All scores are shown for illustration only and do not reflect actual model performance.}
\label{fig:stackingnet}
\end{figure}

This study investigates collective inference among independently trained foundation models, each treated as a black box. Throughout, we use collective inference to denote the post-hoc aggregation of the output predictions of independent, non-interacting models. We propose StackingNet, a lightweight artificial neural network architecture that aggregates only the output predictions of diverse base models. Figure~\ref{fig:stackingnet} presents an overview. The framework unifies regression and classification tasks under a single theoretical and algorithmic framework. Beyond building a meta-combination, StackingNet also enables additional capabilities, including error and group-disparity reduction, reliability ranking, and adversary pruning. Together, these utilities facilitate coordination and robustness among heterogeneous foundation models for future generation AI.

\section*{Results and Discussion}

\subsection*{AI combination outperforms individual AI and matches individual human reviewers}

\begin{figure}[htpb]\centering
\includegraphics[width=1.\textwidth]{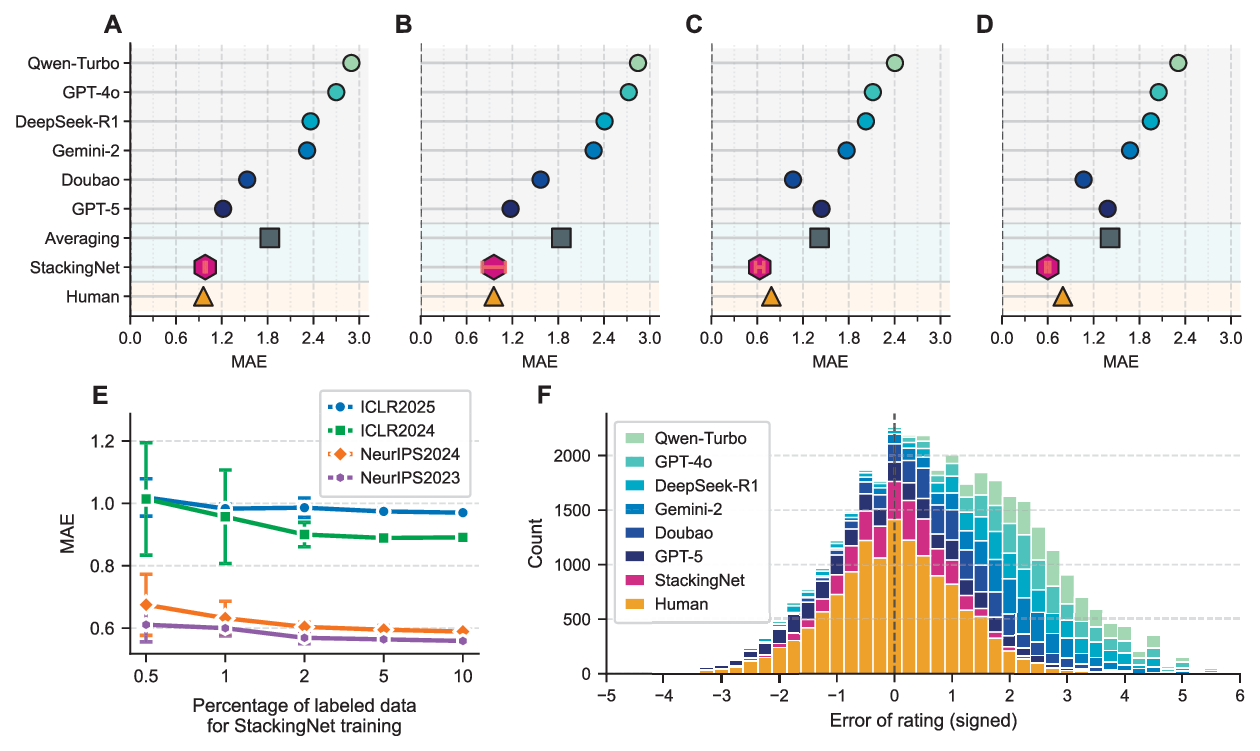}
\caption{{\bfseries Research paper rating error by individual human reviewers, individual LLMs, and collective inference of multiple LLMs.}
{\bfseries a-d}, Mean absolute error (MAE, lower is better) across four datasets: ICLR2025, ICLR2024, NeurIPS2024 and NeurIPS2023. Errors for individual humans are computed relative to consensus scores obtained by aggregating multiple human reviewers. StackingNet was trained in a few-shot setting using 1\% of randomly selected labeled data (10 examples with ground-truth scores).
{\bfseries e}, Performance of StackingNet w.r.t. labeled data size for training.
{\bfseries f}, Distribution of rating errors aggregated across all four datasets.}
\label{fig:paperreview}
\end{figure}

Academic review policies generally prohibit the use of LLMs to generate or substantially draft paper reviews, as such models cannot replace the reading, reasoning and evaluative processes of human experts. To examine this boundary, we compared six proprietary API-based LLMs, individual human reviewers, and combination methods for multiple LLMs across four research paper rating datasets from major AI conferences. Consensus ground-truth scores were obtained by aggregating multiple human ratings per paper. Results are shown in Figure~\ref{fig:paperreview}.

Figure~\ref{fig:paperreview}A-D shows that individual LLMs had higher error than individual human reviewers on all datasets. Without aggregation or alignment, current models are less accurate than human experts in evaluating scientific work. Combining predictions consistently reduced error relative to single models. Even simple averaging outperformed most individual LLMs, while StackingNet achieved the lowest mean absolute error (MAE) on every dataset. These results indicate that aggregation mitigates variance across black-box models and yields more stable inferences.

Aggregating AI predictions matched or exceeded the accuracy of an individual human reviewer. As Figure~\ref{fig:paperreview}A-D shows, StackingNet was better than the average individual human reviewer on every dataset, indicating that collective inference across foundation models can match, and at times improve upon, the reliability of a single expert's numerical rating.

To compare a single human reviewer against an AI ensemble on an equal footing, we further adopt a stricter held-out protocol. For each paper we designate one randomly chosen reviewer as the single-human predictor and build the reference consensus from the remaining reviewers only, a target that reviewer did not contribute to. This is a more conservative design than scoring every predictor against the full consensus, because it holds the single human and every model to the same external quantity. Supporting Information Table~\ref{tab:humanloo} reports this comparison, in which the single human and every model are scored against the held-out consensus with the random choice resampled by bootstrap. Under this strict protocol, a single LLM deviates from the consensus to a degree comparable to a single human reviewer, whereas StackingNet predicts the held-out consensus more accurately than an individual reviewer on all four venues, for example with a mean absolute error of 0.72 versus 1.02 on NeurIPS2023 and 1.12 versus 1.27 on ICLR2025. Its ratings therefore align with the human-expert consensus at least as closely as an individual expert's do. This reflects the variance reduction of aggregation rather than higher-quality judgment, since averaging several imperfect estimates of the same target lowers variance whether the estimators are people or models, making the ensemble a more reliable estimate of where the peer consensus will fall.

Minimal supervision was sufficient to align AI predictions with human ratings. As Figure~\ref{fig:paperreview}E shows, StackingNet is a compact architecture with few learnable parameters that requires only limited labeled data for training. Using 1\% of the data with ratings from human reviewers, equivalent to ten papers as of a few-shot example scenario, was sufficient for high performance of StackingNet. Additional supervision provided only marginal improvements, indicating that minimum supervision is enough to build StackingNet. As Supporting Information Figure~\ref{fig:paperreview-supp} shows, leveraging labeled data from prior years also provides effective supervision and can substantially reduce annotation demands for the current evaluation. These observations suggest a practical paradigm: a small number of annotated samples, with human reviewer ratings for published papers, is sufficient to align the ensemble of LLMs to the preferences of reviewers on the same publication.

Collective inference benefits both humans and AI. Figure~\ref{fig:paperreview}F shows the error distributions, where individual reviewers, whether human or model-based, often diverge from consensus. Most LLMs in the evaluation pool, except GPT-5, tended to assign systematically higher scores. Observe that better-performing models with lower MAE also have less directional divergence. Additionally, note that ratings from individual human reviewers also displayed large variability. Aggregation through StackingNet, in contrast, produced estimates that were more stable. Together, these results illustrate a general principle: aggregating independent predictions brings estimates closer to consensus and reduces the influence of individual variability.

\subsection*{StackingNet reduces individual-model error and group-wise disparities}

Foundation models are increasingly used for data annotation~\cite{Tan2024} and automated judgment~\cite{Li2025LLMjudge}, raising concerns about bias and fairness. Linguistic and visual representations learned from large datasets often encode human-like biases~\cite{Caliskan2017, Liang2021LLMbias}. These biases may arise from multiple sources, including inadequate pre-training data, model architecture, learning objectives, and annotated data. Consequently, LLMs/VLMs differ in both the magnitude and direction of their predictions and errors. By combining predictions across independently developed models, StackingNet can average out errors that are idiosyncratic to individual models.

\begin{figure}[htpb] \centering
\includegraphics[width=.9\textwidth]{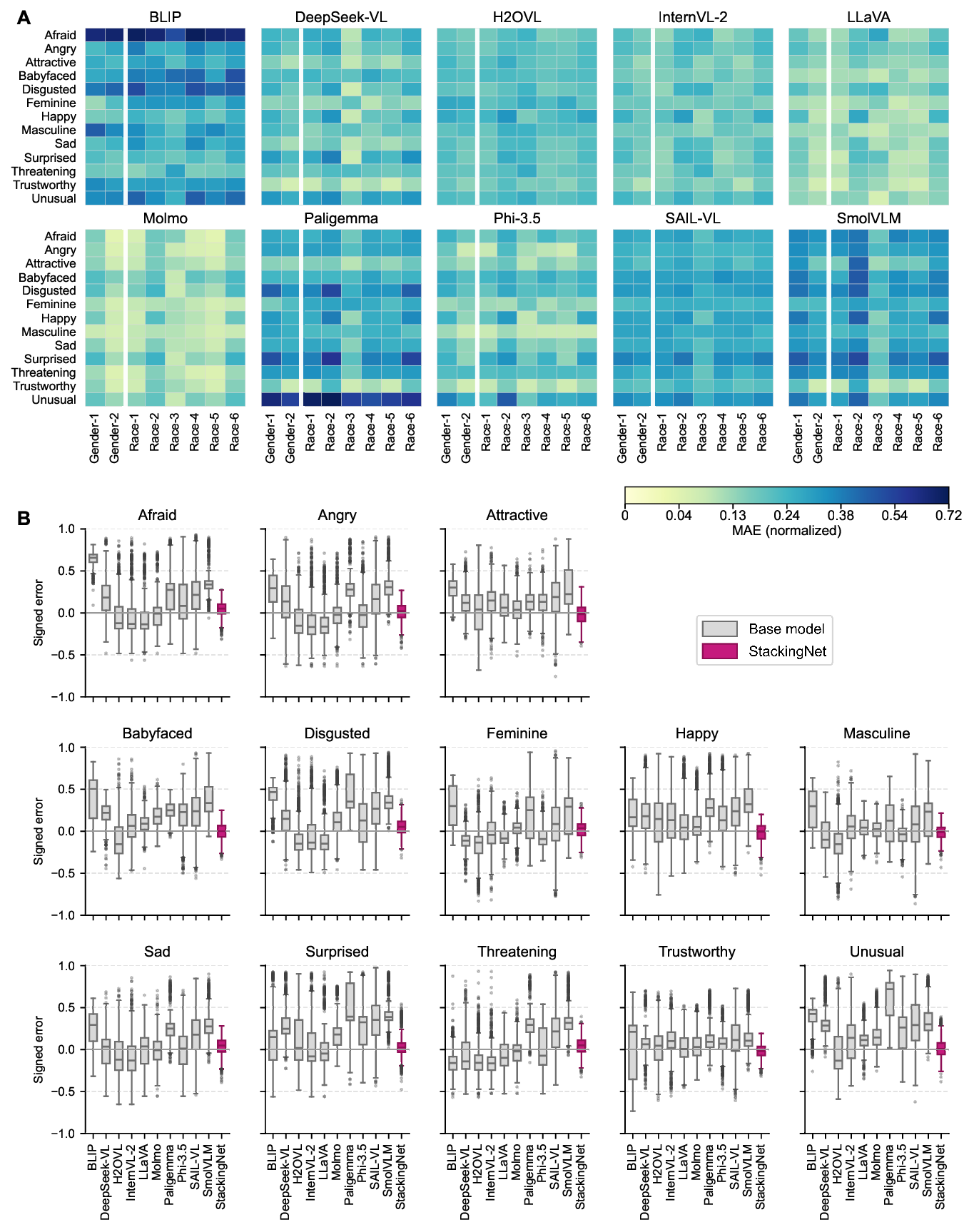}
\caption{{\bfseries Facial attribute ratings by VLMs and StackingNet combination on the Chicago Face Database.}
{\bfseries a}, Heatmaps of MAE of base models across thirteen attributes, stratified by two gender groups and six race/ethnicity groups.
{\bfseries b}, Distribution of signed prediction errors for the base models and StackingNet across thirteen attributes in the normalized label space.
Predictions were clipped, and the label space was min-max normalized to $[0,1]$. Gender and race/ethnicity group names were anonymized with numeric identifiers. Groupings are shown only to audit model behavior and do not imply any inherent traits or conclusions about individuals or populations.}
\label{fig:cfd}
\end{figure}

We evaluated this property on attribute rating tasks using facial images from the Chicago Face Database~\cite{Ma2015}. The database is a standardized set of high-resolution facial photographs, each accompanied by human ratings of perceived attributes. Thirteen attributes were predicted by ten VLMs across two gender groups and six race/ethnicity groups. Figure~\ref{fig:cfd}A summarizes the MAE for each model. Two distinct forms of bias were observed. Absolute bias arises when a model consistently deviates from ground-truth ratings (from human raters) in its overall scoring tendencies, for instance, when some models uniformly assign higher or lower values across all samples. Systematic bias, in contrast, reflects uneven performance across demographic groups. Accordingly, certain models, e.g., H2OVL, achieved lower overall MAE, whereas others, e.g., BLIP, exhibited pronounced demographic disparities. In general, differences across gender were smaller than those across race/ethnicity, although attributes inherently associated with "masculine" or "feminine" still showed measurable bias. These findings reveal substantial variation both between and within models.

To further analyze directional tendencies, we examined signed prediction errors. As shown in Figure~\ref{fig:cfd}B, models such as BLIP and SmolVLM exhibited global overestimation or underestimation, while others produced asymmetric, long-tailed error distributions for specific attributes such as "disgusted" or "surprised". These skewed distributions indicate that even models with better overall performance can display sample-level instability and directional preference. Such inconsistencies may propagate to downstream tasks, introducing systematic biases that are difficult to detect yet are dangerous.

The observed disparities across race/ethnicity likely stem from improper representation in training data. When models are trained on socially sensitive attributes such as "trustworthy", they may inadvertently amplify stereotypes embedded in the training corpora or overcompensate to avoid perceived discrimination. As shown in Supporting Information Figure~\ref{fig:cfd-supp}, VLMs such as Paligemma and SmolVLM substantially overestimated "trustworthy" for the Race-2 and Race-6 groups, reflecting sensitivity to societal biases in pre-training data. In these cases, societal norms surrounding racial discrimination appear to influence model behavior, leading to overly high ratings that, ironically, constitute another form of systematic bias. This tendency was not observed for the other race/ethnicity groups. By contrast, smaller gender gaps likely result from more balanced gender representation in the training data.

Because training data will never be entirely fair or complete, no single proprietary model can be expected to be free of such biases. StackingNet does not reweight data. Instead, it leverages diversity across independently developed models, so that errors specific to any one model are diluted in the aggregate. This helps when models disagree in the direction of their errors, but it offers no protection when they agree. Table~\ref{tab:fairness} shows that for the socially sensitive ``trustworthy'' attribute, for instance, a bias shared across the pool survives aggregation and can even be reinforced.

Beyond these per-model patterns, we asked whether StackingNet's reduction in average error also amounts to fairness across groups. Figure~\ref{fig:cfd}B shows the most direct evidence. Trained with only 1\% labeled data (14 annotations per attribute), StackingNet collapsed the wide, directionally skewed error distributions of the individual models into a single compact distribution centered near zero. The global over- and under-estimation seen in models such as BLIP and SmolVLM and the long attribute-specific tails of the base models are largely absorbed in the aggregate, leaving little residual directional bias. Averaged over attributes, this tightening lowers the worst-group MAE from 0.164 to 0.122 across race/ethnicity and from 0.143 to 0.093 across gender, and roughly halves the largest between-group gap.

To confirm that this reflects genuine group-level improvement and not merely a lower overall average, we examined two standard group-fairness views~\cite{Caton2024, Castelnovo2022}, computed separately across the six race/ethnicity groups and the two gender groups. The first asks whether the error is equally low across groups, a property known as accuracy parity~\cite{Chi2021, Agarwal2019, Diana2021}. Table~\ref{tab:fairness} shows that StackingNet lowers the worst-group MAE on all 13 attributes by gender and on 11 of 13 by race/ethnicity, and that the improvement over the best base model is significant at the bootstrap 95\% level, with the confidence interval lying entirely below zero, on 13 of the gender attributes and 10 of the race/ethnicity attributes. It also narrows the largest between-group gap on 11 of 13 attributes by race/ethnicity.

The second view asks whether the predictions themselves, rather than their errors, are distributed similarly across groups, a property known as demographic parity~\cite{Chzhen2020}. The difference is that demographic parity concerns the absolute predicted scores, whereas accuracy parity concerns the error relative to the human labels. The former measures whether group differences are amplified, and the latter whether the predictions are accurate. Supporting Information Figure~\ref{fig:fairness-dp} reports this, the between-group spread in the mean predicted score, quantified analogously to the accuracy-parity gap as the difference between the largest and smallest per-group mean prediction. By this measure StackingNet does not widen the group differences already present in the human labels. On race/ethnicity its predictions are no more separated across groups than the labels on every attribute, and markedly more balanced on average, with the between-group difference falling from 0.154 in the labels to 0.046, while the gender-defining ``masculine'' and ``feminine'' correctly retain their real group signal.

\begin{table}[htpb] \centering \setlength{\tabcolsep}{1.0mm} \renewcommand{\arraystretch}{1.1}
\caption{Group-fairness statistics on the Chicago Face Database.}
\label{tab:fairness}
\begin{threeparttable}
\resizebox{\textwidth}{!}{\begin{tabular}{l|cccc|cccc}
\toprule
& \multicolumn{4}{c|}{Race/ethnicity (6 groups)} & \multicolumn{4}{c}{Gender (2 groups)} \\
\cmidrule(lr){2-5}\cmidrule(lr){6-9}
& \multicolumn{2}{c}{Worst-group MAE $\downarrow$} & \multicolumn{2}{c|}{Between-group gap $\downarrow$} & \multicolumn{2}{c}{Worst-group MAE $\downarrow$} & \multicolumn{2}{c}{Between-group gap $\downarrow$} \\
Attribute & Best base & StackingNet & Best base & StackingNet & Best base & StackingNet & Best base & StackingNet \\
\midrule
afraid & 0.176 & \textbf{0.102} & 0.134 & \textbf{0.033} & 0.156 & \textbf{0.091} & 0.024 & \textbf{0.005} \\
angry & 0.167 & \textbf{0.091} & 0.115 & \textbf{0.026} & 0.135 & \textbf{0.087} & 0.008 & \textbf{0.002} \\
attractive & 0.162 & \textbf{0.131} & 0.092 & \textbf{0.044} & 0.141 & \textbf{0.118} & \textbf{0.018} & 0.031 \\
babyfaced & 0.144 & \textbf{0.083} & 0.059 & \textbf{0.009} & 0.119 & \textbf{0.080} & \textbf{0.003} & 0.004 \\
disgusted & 0.240 & \textbf{0.089} & 0.164 & \textbf{0.036} & 0.210 & \textbf{0.083} & 0.024 & \textbf{0.004} \\
feminine & 0.111 & \textbf{0.100} & 0.048 & \textbf{0.024} & 0.116 & \textbf{0.092} & 0.054 & \textbf{0.018} \\
happy & \textbf{0.212} & 0.236 & \textbf{0.139} & 0.150 & 0.169 & \textbf{0.118} & \textbf{0.006} & 0.009 \\
masculine & 0.102 & \textbf{0.083} & 0.040 & \textbf{0.015} & 0.105 & \textbf{0.083} & 0.027 & \textbf{0.019} \\
sad & 0.155 & \textbf{0.111} & 0.082 & \textbf{0.027} & 0.143 & \textbf{0.093} & 0.018 & \textbf{0.001} \\
surprised & 0.216 & \textbf{0.159} & 0.138 & \textbf{0.128} & 0.177 & \textbf{0.082} & 0.003 & \textbf{0.001} \\
threatening & 0.150 & \textbf{0.112} & 0.100 & \textbf{0.032} & 0.120 & \textbf{0.093} & \textbf{0.003} & 0.009 \\
trustworthy & \textbf{0.150} & 0.175 & \textbf{0.086} & 0.113 & 0.134 & \textbf{0.108} & \textbf{0.028} & 0.028 \\
unusual & 0.148 & \textbf{0.115} & 0.089 & \textbf{0.066} & 0.139 & \textbf{0.075} & 0.012 & \textbf{0.004} \\
\midrule
Avg. & 0.164 & \textbf{0.122} & 0.099 & \textbf{0.054} & 0.143 & \textbf{0.093} & 0.017 & \textbf{0.010} \\
\bottomrule
\end{tabular}}
\begin{tablenotes}[flushleft]
\footnotesize
\item Worst-group MAE is the maximum per-group mean absolute error.
\item The between-group gap is the worst-group minus best-group MAE.
\item All values are in the normalized $[0,1]$ label space.
\item Lower is better for both statistics.
\item Bold marks the better of the best base model and StackingNet.
\end{tablenotes}
\end{threeparttable}
\end{table}

These reductions call for careful interpretation. Lower group-wise error indicates closer alignment with the human-consensus reference, not the removal of the bias that the reference itself carries. Each Chicago Face Database rating is itself a consensus, averaged over a large and demographically diverse pool of 1{,}087 raters. Each rater judged a random subset of ten to fifteen faces, so that every face was scored by many dozens of raters (on the order of seventy on average), and raters were instructed to judge each face relative to others of the same race/ethnicity and gender~\cite{Ma2015}. This averaging already reduces idiosyncratic individual bias, much as ensembling does, but it also encodes the raters' shared, within-group social norms into the labels. Attributes such as ``trustworthy'' or ``threatening'' are over-generalized social judgments rather than objective facial properties~\cite{Oosterhof2008, Todorov2015, Caliskan2017}, so improving accuracy with respect to them can reproduce, rather than remove, culturally shared stereotypes. The statistical disparities StackingNet reduces are therefore measured relative to this reference, whose social bias resides in the data and the labels rather than in the aggregation method.

\subsection*{StackingNet unifies regression and classification within a single framework}

As illustrated in Figure~\ref{fig:stackingnet}B, beyond continuous rating tasks, the proposed StackingNet also addresses classification problems involving discrete categorical decisions, including binary and multi-class classification tasks. Eight datasets from the HELM benchmark~\cite{Liang2023} were evaluated, covering tasks such as true-false judgment, yes-no questions, sentiment classification, and logical reasoning. Ten API-based LLMs were queried to generate predictions. Table~\ref{tab:classification} shows the results for prediction aggregation across these models.

\begin{table}[htpb] \centering \setlength{\tabcolsep}{1.1mm}
\caption{Balanced classification accuracy (\%) and average ranking of combination algorithms on the classification datasets.}
\label{tab:classification}
\begin{threeparttable}
\resizebox{\textwidth}{!}{\begin{tabular}{l|llllllll|c}
\toprule
Approach & BoolQ & CivilC. & EntityM. & IMDB & LegalS. & LSAT & MMLU & RAFT & Rank \\
\midrule
Base-Worst & 64.13 & 50.91 & 61.74 & 93.37 & 53.77 & 19.07 & 35.51 & 76.77 & - \\
Base-Best & 89.21 & 65.12 & 95.76 & 96.25 & 65.26 & 24.38 & 59.48 & 88.58 & - \\
\midrule
Voting & 87.29$_{\pm0.19}$ & 64.39$_{\pm0.09}$ & 92.01$_{\pm0.64}$ & \textbf{97.17}$_{\pm0.07}$ & 64.64$_{\pm0.46}$ & 21.59$_{\pm0.82}$ & 54.60$_{\pm0.25}$ & 88.06$_{\pm0.30}$ & 13 \\
WAwA & 88.26$_{\pm0.00}$ & 64.79$_{\pm0.00}$ & 93.95$_{\pm0.00}$ & 96.87$_{\pm0.00}$ & 65.00$_{\pm0.00}$ & 21.44$_{\pm0.00}$ & 54.82$_{\pm0.00}$ & 88.10$_{\pm0.00}$ & 5 \\
Dawid-Skene & 88.01$_{\pm0.00}$ & 65.24$_{\pm0.00}$ & 90.33$_{\pm0.00}$ & 96.87$_{\pm0.00}$ & 64.80$_{\pm0.00}$ & \textbf{21.90}$_{\pm0.00}$ & 53.90$_{\pm0.00}$ & 89.62$_{\pm0.00}$ & 9 \\
M-MSR & 88.41$_{\pm0.00}$ & 64.98$_{\pm0.00}$ & 92.86$_{\pm0.00}$ & 96.99$_{\pm0.00}$ & \textbf{65.12}$_{\pm0.00}$ & 19.28$_{\pm0.00}$ & 54.58$_{\pm0.00}$ & 87.17$_{\pm0.00}$ & 6 \\
MACE & 87.74$_{\pm0.00}$ & 64.13$_{\pm0.00}$ & 91.83$_{\pm0.00}$ & 96.99$_{\pm0.00}$ & 63.74$_{\pm0.00}$ & 19.81$_{\pm0.00}$ & \textbf{55.10}$_{\pm0.00}$ & \textbf{89.75}$_{\pm0.00}$ & 11 \\
GLAD & 88.23$_{\pm0.00}$ & 64.92$_{\pm0.00}$ & 94.70$_{\pm0.00}$ & 96.99$_{\pm0.00}$ & 64.37$_{\pm0.00}$ & 19.76$_{\pm0.00}$ & 54.74$_{\pm0.00}$ & 87.30$_{\pm0.00}$ & 10 \\
KOS & 88.24$_{\pm0.00}$ & 64.84$_{\pm0.00}$ & 92.96$_{\pm0.00}$ & 96.99$_{\pm0.00}$ & 65.00$_{\pm0.00}$ & \text{N/A} & \text{N/A} & 87.17$_{\pm0.00}$ & 4 \\
SML & 88.24$_{\pm0.00}$ & 64.84$_{\pm0.00}$ & 89.91$_{\pm0.00}$ & 96.99$_{\pm0.00}$ & 64.94$_{\pm0.00}$ & 21.59$_{\pm0.00}$ & 54.90$_{\pm0.00}$ & 89.05$_{\pm0.00}$ & 2 \\
LA & 88.25$_{\pm0.01}$ & 64.86$_{\pm0.03}$ & 93.25$_{\pm0.00}$ & 96.94$_{\pm0.07}$ & 65.07$_{\pm0.07}$ & 19.98$_{\pm0.05}$ & 54.97$_{\pm0.00}$ & 87.17$_{\pm0.13}$ & 7 \\
LAA & 87.73$_{\pm0.09}$ & 65.46$_{\pm0.02}$ & 90.20$_{\pm0.03}$ & 65.89$_{\pm24.43}$ & 51.77$_{\pm4.44}$ & 20.55$_{\pm1.98}$ & 54.98$_{\pm0.01}$ & 89.56$_{\pm0.11}$ & 14 \\
EBCC & 88.26$_{\pm1.31}$ & 65.21$_{\pm0.10}$ & 88.17$_{\pm1.26}$ & 96.99$_{\pm0.00}$ & 64.58$_{\pm0.51}$ & 21.11$_{\pm0.85}$ & 54.97$_{\pm0.40}$ & 88.56$_{\pm0.70}$ & 3 \\
PM & 88.76$_{\pm0.00}$ & 64.30$_{\pm0.00}$ & 95.19$_{\pm0.00}$ & 96.84$_{\pm0.06}$ & 64.77$_{\pm0.00}$ & 20.10$_{\pm0.00}$ & 54.99$_{\pm0.04}$ & 87.13$_{\pm0.53}$ & 11 \\
ZenCrowd & 88.16$_{\pm0.00}$ & 64.75$_{\pm0.00}$ & 94.41$_{\pm0.00}$ & 96.99$_{\pm0.00}$ & 64.88$_{\pm0.00}$ & 19.72$_{\pm0.00}$ & 54.97$_{\pm0.00}$ & 87.30$_{\pm0.00}$ & 8 \\
U-StackingNet (ours) & \textbf{89.24}$_{\pm0.00}$ & \textbf{65.80}$_{\pm0.00}$ & \textbf{95.22}$_{\pm0.00}$ & 96.97$_{\pm0.05}$ & 64.94$_{\pm0.07}$ & 21.38$_{\pm0.29}$ & 54.42$_{\pm0.04}$ & 88.33$_{\pm0.06}$ & \textbf{1} \\
\midrule
Logistic Regression & 89.13$_{\pm0.32}$ & 50.00$_{\pm0.00}$ & 91.71$_{\pm0.96}$ & \textbf{97.59}$_{\pm0.50}$ & 63.14$_{\pm1.71}$ & \textbf{22.62}$_{\pm1.55}$ & 57.73$_{\pm1.38}$ & 89.11$_{\pm0.64}$ & 2 \\
S-StackingNet (ours) & \textbf{89.75}$_{\pm0.09}$ & \textbf{65.81}$_{\pm0.00}$ & \textbf{96.21}$_{\pm0.00}$ & 97.09$_{\pm0.06}$ & \textbf{65.08}$_{\pm0.11}$ & 19.85$_{\pm1.70}$ & \textbf{58.80}$_{\pm0.42}$ & \textbf{89.18}$_{\pm0.10}$ & \textbf{1} \\
\bottomrule
\end{tabular}}
\begin{tablenotes}[flushleft]
\footnotesize
\item Entries report mean $\pm$ standard deviation over five runs with different random seeds.
\item The worst and best single base models are provided for reference.
\item The rightmost column reports the average rank across datasets (lower is better).
\item For each dataset, the best-performing combination method in unsupervised and supervised learning is highlighted in bold.
\end{tablenotes}
\end{threeparttable}
\end{table}

The results in Table~\ref{tab:classification} show that combination methods, including the unsupervised variant of StackingNet (U-StackingNet), effectively integrate predictions from multiple base models to form a meta-combination. Without any labeled data, these combinations consistently outperformed the weakest base models, and in many cases even surpassed the single best model. This capability enables reliable aggregation even when no ground-truth annotations are available. StackingNet also exceeded the performance of classic and recent combination methods from the crowdsourcing literature~\cite{Li2016}, and can further incorporate limited labeled data to enhance performance beyond.

The key advantage of StackingNet is in its ability to learn jointly from labeled and unlabeled data. When another 10\% of the samples were labeled, the supervised variant (S-StackingNet) used these examples to achieve additional performance gains with learning objectives on both the labeled and unlabeled data. This flexibility distinguishes StackingNet from classic combination methods, which rely solely on supervised or unsupervised aggregation. The only exception occurred in the LSAT dataset, where all base models performed close to random, leaving little room for effective combination. Such results demonstrate that StackingNet is able to handle diverse task types across both subjective regression and objective classification tasks. StackingNet not only outperformed individual base models but also achieved performance gain over state-of-the-art combination approaches in building the meta-ensemble.

\begin{figure}[htpb]\centering
\includegraphics[width=\textwidth]{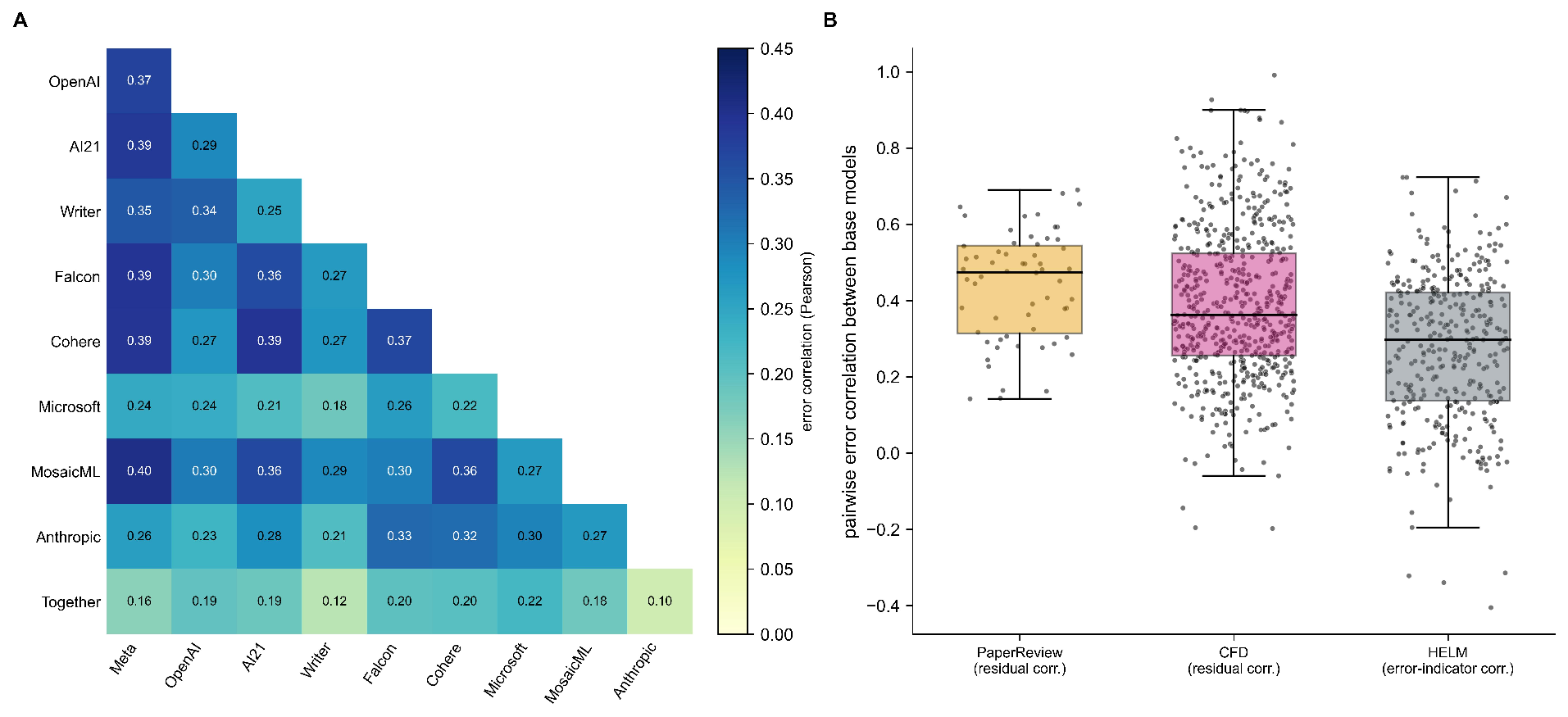}
\caption{{\bfseries Pairwise error correlation between base models.}
{\bfseries a}, Mean error correlation between the ten HELM base models, averaged over the eight HELM classification datasets and shown as the lower triangle of the symmetric matrix.
{\bfseries b}, Distribution of the pairwise error correlations across tasks, grouped into the Chicago Face Database regression attributes, the paper-rating regression venues, and the HELM classification datasets. Boxes show the quartiles and whiskers extend to 1.5 times the interquartile range, and each point is one base-model pair, so points beyond the quartiles fall outside the box.
For the regression tasks, the Pearson correlation between the two models' residuals (the signed prediction errors) was used. For the classification tasks, the Pearson correlation between their per-item error indicators (one on a wrong answer, zero on a correct one) was used. The two error scales are compared only qualitatively.}
\label{fig:helm-corr}
\end{figure}

The effectiveness of such combination depends on how the base models' errors relate to one another. Although StackingNet treats each model as an independent black box, in practice their errors are correlated, because contemporary foundation models share training corpora, architectures, and alignment pipelines, and are frequently trained by distilling knowledge from the outputs of other models~\cite{Hinton2015, Xu2024}. Figure~\ref{fig:helm-corr}A shows that even the diverse HELM pool, which draws one model from each of ten organizations, has a positive error correlation between every pair of models, averaged across the eight HELM datasets, and Figure~\ref{fig:helm-corr}B shows that in every task group the distribution of pairwise correlations sits well above the zero value expected under independence. Supporting Information Table~\ref{tab:dependence} quantifies this for every task, where residual-error correlations average about 0.4, alongside the joint-error rate, the share of items two classifiers get wrong together (the double-fault diversity measure of ref~\cite{Kuncheva2003}), which separates a genuinely redundant pool from a strong pool whose errors merely covary. Because the joint-error rate accounts for accuracy, the two columns need not move together: IMDB, for example, shows a higher error correlation than LSAT yet a far lower joint-error rate, because its base models are individually strong and seldom wrong on the same item, whereas the LSAT models are near chance and fail together often.

\subsection*{StackingNet enables unsupervised ranking for model performance benchmarking}

Accurately ranking the performance of black-box models on downstream tasks is essential for both developers and users, since no single model performs best across all tasks~\cite{Liang2023, Chang2024}. Yet model evaluation on unseen data remains challenging when annotations are scarce. Public benchmarks often fail to represent real-world performance because many widely used evaluation datasets are partially included in model pretraining or fine-tuning. For example, contamination analyses of LLaMA-2~\cite{Touvron2023} and GPT-4~\cite{Achiam2023} revealed that benchmark datasets such as MMLU appear as paraphrased or near-duplicate forms in the pretraining data of LLMs. Such leakage undermines benchmark validity and highlights the need for alternative evaluation strategies.

\begin{figure}[htpb]\centering
\includegraphics[width=1.\textwidth]{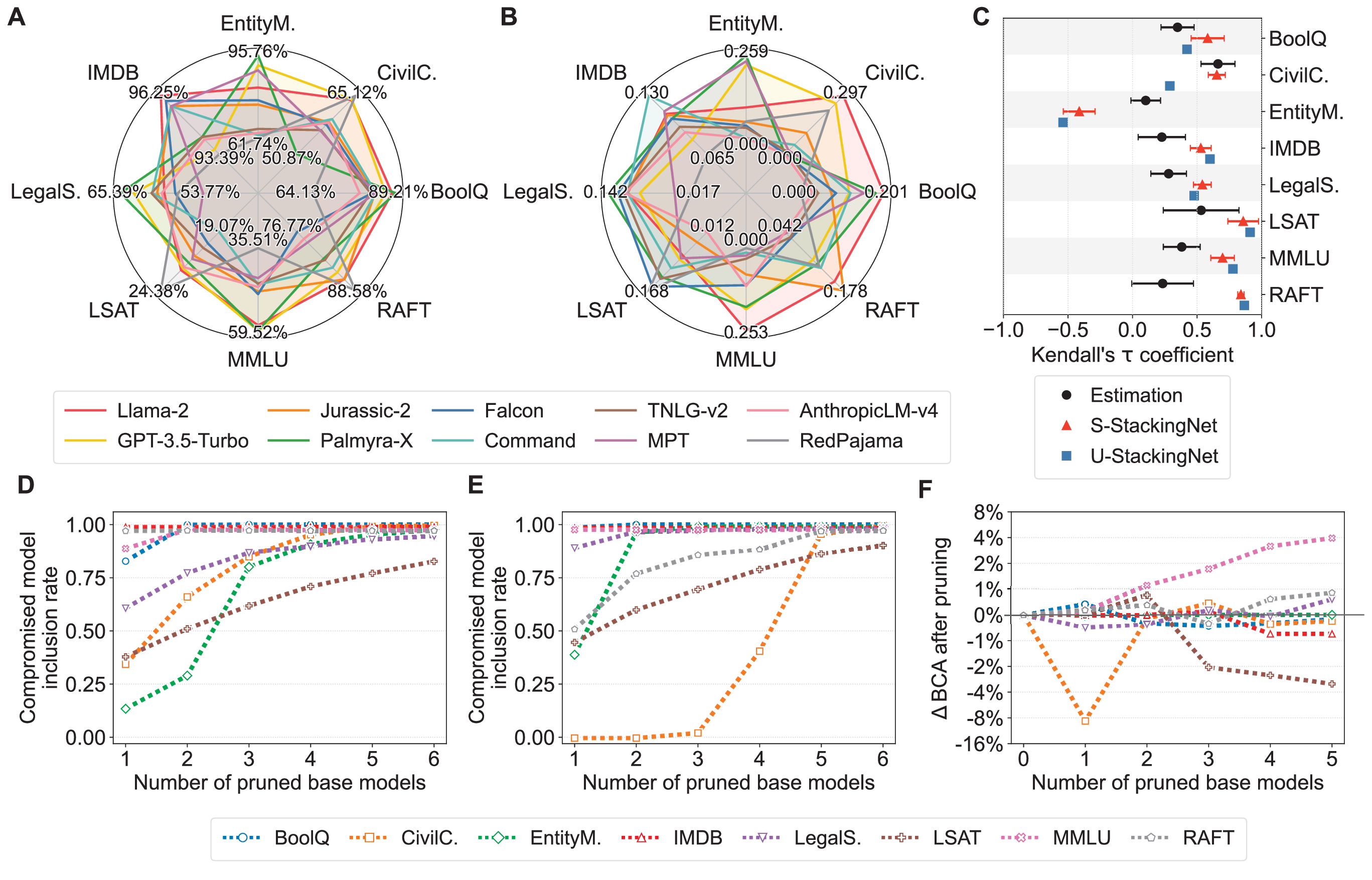}
\caption{{\bfseries Performance of StackingNet on classification tasks for ranking and pruning.}
Panels a-c concern model ranking, and panels d-f concern adversary detection and pruning.
{\bfseries a}, Ground-truth balanced classification accuracy (BCA) of base models.
{\bfseries b}, Learned combination weights of StackingNet, estimated using few-shot annotations. Maximum and minimum values are marked for each dataset in the radar plots.
{\bfseries c}, Kendall’s $\tau$ rank correlation with ground-truth rankings, where higher is better, under few-shot estimation and StackingNet in supervised (S-StackingNet) or unsupervised (U-StackingNet) settings.
Results represent the average of five runs with different random seeds.
{\bfseries d}, Detection of random injection attacks by StackingNet after inserting a randomly predicting base model.
{\bfseries e}, Detection of label flipping attacks after flipping predictions from the best-performing base model.
{\bfseries f}, Change in combination performance after sequential pruning of the lowest-performing base model using StackingNet's trained weights.}
\label{fig:rankprune}
\end{figure}

StackingNet addresses this challenge by enabling reliable model ranking with minimal or no labeled data. Figure~\ref{fig:rankprune}A shows the ground-truth performance of ten API-based LLMs on eight HELM benchmark datasets. Figure~\ref{fig:rankprune}B presents rankings estimated by StackingNet using ten labeled samples per class, simulating a few-shot scenario where only very limited samples have labels. The radar plots show that the estimated model ordering closely matches the ground-truth ranking over the entire test set. Observe that StackingNet assigns higher weights to better-performing base models and near-zero weights to weaker ones in the combination. Figure~\ref{fig:rankprune}C shows that, in contrast, direct few-shot estimation based only on limited annotations yields unstable results with large variance. StackingNet achieved over 0.5 Kendall’s $\tau$ correlation on seven datasets, indicating reliable ranking performance under few-shot supervision. The exception observed for the EntityMatching dataset arises because most base models perform poorly on a similar scale, making reliable ranking difficult.

Figure~\ref{fig:rankprune}C shows that, even without labeled data, the unsupervised U-StackingNet estimated reliability weights that strongly correlate with ground-truth performance. Its inferred rankings closely followed the true ordering across all eight datasets and substantially outperformed direct few-shot estimation with annotations. By exploiting inter-model correlations, StackingNet recovers underlying reliability and enables interpretable ranking when annotations are expensive to query. 

This utility of StackingNet also has broad implications for benchmarking practice. Traditional benchmarks publicly release labeled test sets, exposing them to contamination and diminishing their long-term validity. By contrast, StackingNet that operates with limited or no annotations allow performance comparison without label disclosure, reducing the risk of data leakage while maintaining evaluative power. Such strategies could support a more secure and sustainable benchmarking paradigm, preserving the integrity of evaluation datasets as AI systems continue to evolve.

\subsection*{StackingNet enables robust pruning against compromised base models}

Beyond ranking, another important utility of StackingNet is adversary pruning. With the rise of middleware platforms, also known as model routers, that host and route queries across such APIs~\cite{Ong2025}, users often interact indirectly with foundation models through intermediaries. Although these platforms simplify access and provide a natural setting for building StackingNet, they also introduce vulnerabilities such as connection failures and, more importantly, input/response manipulation by attackers~\cite{Zhao2024a}.

StackingNet architecture provides an inherent mechanism for detecting and pruning compromised models within the ensemble. Two attack scenarios were evaluated: random injection and label flipping. In random-injection attacks, a new base model is inserted that outputs random predictions. In label-flipping attacks, the predictions of the best-performing model are intentionally inverted to produce incorrect outputs. Such attacks are easy to implement and do not require access to information from other base models in the ensemble.

StackingNet detects these compromised models through ranking their learned weight. As shown in Figure~\ref{fig:rankprune}B, the framework assigns substantially lower weights to corrupted models after training, thereby enabling identification of the base model with the lowest weight as highly compromised. In Figure~\ref{fig:rankprune}D-E, when half of the models in the ensemble are inspected, the detection rate almost converges to one on all datasets. These findings demonstrate that StackingNet can identify and isolate compromised contributors even when adversary is injected into the combination.

StackingNet also enables pruning, in which unreliable base model with the lowest weight is removed to improve ensemble robustness. Figure~\ref{fig:rankprune}F shows the results of iteratively removing the lowest-performing model and retraining StackingNet across eight datasets. Pruning half of the weaker models led to either improved or unchanged performance in nearly all cases. The only exception was the LSAT dataset, where the absolute performance of all base models was low. Supporting Information Figure~\ref{fig:attackprune-supp} provides additional pruning results when base models are compromised. These results confirm that StackingNet supports effective model pruning and maintains stable ensemble performance even under potential adversarial interference. Notably, this pruning capability also delivers practical utility when the number of base models is large: by eliminating redundant or low-contribution models, it directly reduces the query costs associated with ensemble inference while maintaining ensemble performance.

\subsection*{Discussion}

StackingNet's unsupervised aggregation rests on a conditional-independence assumption (Methods, Assumption~\ref{assumption:independence}) that is an idealization in practice, since the base models' errors are in fact correlated rather than independent. The supervised StackingNet, however, assumes none of this. Fit by empirical risk minimization on labeled data, it estimates its weights against the observed error structure and so down-weights correlated, redundant models rather than presuming independence. This robustness is visible in two experiments. Supporting Information Figure~\ref{fig:degradation} shows that, in a controlled synthetic setting, growing a cluster of perfectly correlated models drives unweighted averaging to the single-model error level while StackingNet stays well below it. With real models, Supporting Information Figure~\ref{fig:helm-robust} shows that adding redundant model variants from a single organization to a diverse pool drags averaging down while StackingNet maintains accuracy across four organizations and two datasets. Complete conditional independence is therefore hard to achieve in practice, yet StackingNet continues to help even when it does not hold.

A central question in ensemble learning is whether combining models truly yields more reliable predictions than selecting the single best performer~\cite{Dzeroski2004, Li2025SML-OVR}. Although traditional evaluation pipelines emphasize identifying a single winning model, this strategy is unstable. As Table~\ref{tab:classification} and Figure~\ref{fig:rankprune} show, single-model selection depends on uncertain prior estimates of performance and varies across datasets. Theorem~\ref{theorem:votingconvergence} formalizes that ensemble accuracy converges upward as competent learners are added, whereas the performance of any individual model remains bounded by its own bias and variance, and empirically aggregation consistently outperforms even the best standalone model. Pruning weak models does not always help, however. As Figure~\ref{fig:rankprune}F shows, apparent underperformers may still carry complementary information that improves collective generalization, and identifying which models underperform is itself uncertain. Diversity is therefore not a byproduct but a prerequisite for robust collective inference.

Where do these gains originate? Supporting Information Figure~\ref{fig:dispersiongain} plots, for each task, how much StackingNet beats unweighted averaging, the simplest combiner available without labels, against the dispersion of the base models, the quality gap between the best and worst model in the pool. Each point is one task, its horizontal position measuring how unequal the base models are and its vertical position measuring StackingNet's gain over averaging, so a value above the dashed zero line marks a task where StackingNet outperforms averaging and a higher point is a larger margin. Read this way, the cloud of points slopes upward, and the gain grows with the dispersion of the pool on both task types, with a rank correlation coefficient of $+0.62$ ($p=0.03$) on the Chicago Face Database and $+0.81$ ($p=0.02$) across the HELM datasets. This is the signature of the underlying mechanism: when base models differ in quality, averaging is pulled toward the weaker members while StackingNet learns to down-weight them, so its advantage widens as the quality gap grows. When the pool is uniformly strong, in contrast, the advantage is small and occasionally slightly negative, as on the low-dispersion IMDB and LSAT tasks, because averaging is already near-optimal there and little is left to gain over it.

This is also why we measure the gain against averaging rather than against the single best base model: the best base cannot be identified in advance without labels, whereas averaging is always available. Even when the pool is restricted to top-ranked, closely matched models, StackingNet remains ahead of the best single model on the paper-rating tasks, a residual benefit from combining strong but imperfectly correlated predictors. The paper-rating and classification experiments draw on different model pools, proprietary LLMs in the first and the public HELM release in the second, which provides complementary evidence across task types rather than a single cross-domain comparison.

Collective inference, in the sense studied here, is distinct from collective intelligence. The gains we document arise from variance reduction and consensus alignment among independent, non-interacting models. They do not arise from the inter-agent interaction, feedback, or emergent group-level cognition studied in the interdisciplinary collective-intelligence literature. The distinction matters for interpretation, because StackingNet coordinates models that never communicate, so its benefits are statistical rather than the product of any group reasoning. This independence is itself an advantage. Aggregating the separate judgments of non-interacting models can outperform organizing them to deliberate together, because withholding interaction preserves the individual reasoning and uncorrelated errors on which effective aggregation depends.

Our paper-rating results bear on a sensitive question: can an ensemble of models match or surpass a human reviewer? As Supporting Information Table~\ref{tab:humanloo} reports and Figure~\ref{fig:paperreview} illustrates, under a strict protocol that scores a single reviewer and the models against the same held-out consensus of the remaining reviewers, StackingNet predicts that consensus more accurately than an individual reviewer on all four venues. This reflects the variance reduction of aggregation rather than superior scientific judgment, because averaging several imperfect estimates of the same target lowers variance whether the estimators are people or models, so the ensemble is a more reliable estimate of where the peer consensus will fall.

The same logic applies to human peer review itself. Because aggregation lowers variance only when it combines informative estimates that are not perfectly correlated, a panel of more reviewers from different institutions and backgrounds, whose judgments are less likely to share the same idiosyncratic bias, should yield a more reliable consensus than a small or homogeneous one. Figure~\ref{fig:paperreview}F shows the individual-reviewer variability that such aggregation reduces, and Figure~\ref{fig:helm-corr} shows the residual cross-organization error correlation, indicating that diversity of perspective, not panel size alone, is what limits shared bias.

Our paper-rating evaluation, moreover, draws on publicly accessible papers from major AI conferences, whose subject matter the language models have likely encountered during training, so these results need not generalize to the full breadth of academic research. Quantitative ratings capture only one facet of scientific quality, and the interpretative and creative reasoning of expert review remains beyond what a score can represent. Collective AI ratings should therefore serve as a reference that supports human evaluation, helping to flag inconsistency and bias, rather than as a substitute for it. As AI systems take on more assessment, transparency about how such scores are produced will be essential for accountability~\cite{Tang2024}.

The fairness results require careful interpretation. As Table~\ref{tab:fairness} shows, StackingNet lowers the worst-group MAE on the large majority of attributes, and Supporting Information Figure~\ref{fig:fairness-dp} shows that its predictions are left no more separated across groups than the human labels. Accuracy with respect to socially sensitive attributes, however, is not the same as fairness in a deeper, social sense. The reference ratings are a human consensus that, by design, was elicited relative to others of the same race/ethnicity and gender~\cite{Ma2015}, and such face-trait judgments are over-generalized social attributions that can encode culturally shared stereotypes~\cite{Oosterhof2008, Todorov2015, Caliskan2017}. Improving agreement with these labels may reproduce, rather than eliminate, those stereotypes. The ``trustworthy'' attribute illustrates the risk directly, as it is the one socially sensitive trait whose worst-group error StackingNet does not reduce. An aggregation method that operates only on model outputs can measure and reduce disparities relative to this reference, but it cannot correct the bias carried by the reference itself.

StackingNet’s architecture is well suited for deployment as a middleware, or model router, on platforms that coordinate multiple API-based foundation models. Currently, such platforms can already route user queries to multiple proprietary systems, and face challenges of ranking and pruning. With its interpretable and lightweight design, StackingNet offers a secure and efficient approach to handle such problems. In this context, collective inference functions not only as a computational strategy but also as a governance tool, enabling effective, explainable, and robust coordination across diverse AI foundation models. Figure~\ref{fig:stackingnet}D summarizes these utilities.

As Supporting Information Figures~\ref{fig:sensitivity} and~\ref{fig:weight-sensitivity} show, StackingNet is computationally efficient: its training converges within seconds and its performance remains stable across hyperparameter settings. This contrasts with the vast computational demands of foundation models, underscoring that sophisticated coordination does not require large-scale retraining but rather principled aggregation. Such efficiency suggests feasibility for real-world deployment in AI service platforms where ensemble inference can operate seamlessly with minimal overhead.

\section*{Conclusions}

This work highlights a shift in AI research from improving isolated models to coordinating collective behavior among multiple independent systems. As foundation models become increasingly specialized and independent, coordination will determine the next frontier of intelligence, transforming model diversity from a limitation into an advantage for effectiveness, fairness, and robustness. StackingNet showcases how aggregation can convert distributed knowledge into coherent, trustworthy inference, marking a step toward collaborative AI infrastructures that mirror the distributed nature of human scientific and societal intelligence. 

\section*{Methods}\label{methods}

\subsection*{Assumptions}

Four assumptions are required for the theoretical derivations for inference combination in ensemble learning. Among them, two are fundamental and necessary for both regression and classification, and they are also essential for the effectiveness of StackingNet.

The first is the classic i.i.d. assumption:
\begin{assumption} [Test data i.i.d.] \label{assumption:iid}
Test samples are independently and identically distributed under the unknown test data distribution $\mathrm{D}$:
\begin{align}
p(\mathbf{x}_1, \ldots, \mathbf{x}_n) = \prod_{i=1}^n p(\mathbf{x}_i),
\qquad \mathbf{x} \sim \mathrm{D}.
\end{align}
\end{assumption}

The second is the core assumption for combination method in ensemble learning. It requires the predictions to be independent across base learners on the test set.

\begin{assumption} [Conditional independence] \label{assumption:independence}
The base learners $h(\mathbf{x})$ are conditionally independent given the true label:
\begin{align}
\mathbb{P} \left( h_1(\mathbf{x}), h_2(\mathbf{x}), \ldots, h_M(\mathbf{x}) \mid y \right)
= \prod_{j=1}^{M} \mathbb{P} \left( h_j(\mathbf{x}) \mid y \right),
\qquad (\mathbf{x},y) \sim \mathrm{D}. \label{eq:conditionalindependence}
\end{align}
\end{assumption}

The third assumption is for a more relaxed correlation across base regressors in regression combination.

\begin{assumption}[Zero-mean and uncorrelated regression errors]\label{assumption:zeromean-uncorr}
For regression, error terms $\epsilon(\mathbf{x})$ have zero mean and are uncorrelated under the test distribution:
\begin{align}
\mathbb{E}[\epsilon_j(\mathbf{x})] &= 0,\qquad \forall j,
\\
\mathbb{E} \left[\epsilon_j(\mathbf{x}) \epsilon_{j'}(\mathbf{x})\right] &= 0,\qquad \forall j\neq j'.
\label{eq:uncorrelatederror}
\end{align}
\end{assumption}

The fourth is an additional necessary assumption for classification combination.

\begin{assumption} [Major competence] \label{assumption:majorcompetence}
A majority of classifiers outperform random guessing, i.e., balanced classification accuracy $\pi$ surpasses chance-level prediction:
\begin{align}
\sum_{j=1}^{M} \mathbb{I}\left(\pi_j > \frac{1}{K}\right) > \frac{M}{2},
\end{align}
\end{assumption}

\noindent
We emphasize the following points regarding these assumptions:
\begin{enumerate}
\item \textbf{Assumption~\ref{assumption:iid}} is generally satisfied when the test samples are non-repetitive and independently drawn.
\item \textbf{Assumption~\ref{assumption:independence}} requires that the base models produce sufficiently diverse and conditionally independent predictions.
\item \textbf{Assumption~\ref{assumption:zeromean-uncorr}} is strong and rarely holds exactly in practice.
\item \textbf{Assumption~\ref{assumption:majorcompetence}} is mild and holds for most practical ensembles, failing only when a substantial fraction of models are compromised.
\end{enumerate}

\subsection*{Theoretical foundation for regression combination}

StackingNet derives its architectural design, learning objective, and parameter constraints from the theoretical principles of ensemble learning~\cite{Zhou2012}. The following theorems establish the foundations for uniform-weight and weighted combinations, and the conditions under which optimal weights can be analytically obtained.

\subsubsection*{Uniform-weight combination}

We first examine whether combining $M$ base regressors $\{h_j\}_{j=1}^{M}$ can improve performance over individual models. The uniform-weight combination for a sample $\mathbf{x}$ is defined as
\begin{align}
\hat{h}(\mathbf{x}_i) := \frac{1}{M} \sum_{j=1}^{M} h_j(\mathbf{x}_i).
\end{align}

\begin{theorem}[Error bound for uniform-weight regression combination \cite{Breiman1996}] \label{theorem:regression-theory}
Under Assumptions~\ref{assumption:iid} and \ref{assumption:independence}, the expected squared error of the uniform-weight combination satisfies
\begin{align}
\mathbb{E}\left[(\hat{h}(\mathbf{x}) - y)^2\right] \ \leq \ \frac{1}{M} \sum_{j=1}^{M} \mathbb{E}\left[(h_j(\mathbf{x}) - y)^2\right],
\end{align}
where $y$ is the ground-truth label, and the expectation is taken over $(\mathbf{x},y) \sim \mathrm{D}$, the unknown test data distribution.
\end{theorem}

\noindent
If, in addition, the errors of the base regressors have zero mean and are mutually uncorrelated, the exact error reduction factor can be derived.

\begin{lemma}[Simplified error reduction factor for uniform-weight regression combination \cite{Zhou2012}] \label{lemma:reduction-factor}
Let the mean individual error be
\begin{align}
\overline{\mathrm{err}}(h)
:= \frac{1}{M}\sum_{j=1}^{M}
\mathbb{E}\!\left[(h_j(\mathbf{x}) - y)^2\right].
\end{align}
Under Assumptions~\ref{assumption:iid}, \ref{assumption:independence}, and~\ref{assumption:zeromean-uncorr}, the uniform-weight regression combination's expected squared error satisfies
\begin{align}
\mathrm{err}(\hat{h})
= \frac{1}{M}\,\overline{\mathrm{err}}(h),
\end{align}
Hence, the error of the ensemble is reduced by a factor of $M$ compared with the average individual error.
\end{lemma}

\subsubsection*{Weighted combination}

A more general form of weighted combination introduces non-uniform weights that satisfy the convex combination constraint:
\begin{align}
\widehat{H}(\mathbf{x}_i) = \sum_{j=1}^{M} w_j\, h_j(\mathbf{x}_i), \qquad w_j \ge 0, \qquad \sum_{j=1}^{M} w_j = 1.
\label{eq:weighted-def}
\end{align}

The expected squared error of the weighted combination is
\begin{align}
\mathrm{err}(\widehat{H}) = \sum_{j=1}^M \sum_{j'=1}^M w_j w_{j'} C_{j j'}, \label{eq:convexcomb}
\end{align}
where $C \in \mathbb{R}^{M \times M}$ denotes the error covariance matrix defined as
\begin{align}
C_{j j'} = \mathbb{E}\left[\epsilon_j(\mathbf{x}) \epsilon_{j'}(\mathbf{x})\right],
\end{align}
with $\epsilon_j(\mathbf{x}) = h_j(\mathbf{x}) - y$.
The diagonal entries $C_{jj}$ represent the individual error variances, whereas the off-diagonal entries $C_{jj'}$ capture pairwise error correlations.

\begin{theorem}[Optimal weights for weighted regression combination \cite{Perrone1995}] \label{theorem:weighted-regression}
Under Assumptions~\ref{assumption:iid} and~\ref{assumption:independence}, the optimal weights for weighted combination are
\begin{align}
w_j = \frac{\sum_{j'=1}^M (C^{-1})_{jj'}}{\sum_{j''=1}^M \sum_{j'=1}^M (C^{-1})_{j'' j'}}.
\end{align}
\end{theorem}

\noindent
When the errors of the base regressors are zero-mean and mutually uncorrelated, the optimal solution simplifies by substituting $C_{jj'} = 0$ for all $j \neq j'$.

\begin{lemma}[Closed-form optimal weights and error for weighted regression combination \cite{Perrone1995}] \label{lemma:idealweights}
Under Assumptions~\ref{assumption:iid}, \ref{assumption:independence}, and~\ref{assumption:zeromean-uncorr}, the error of the weighted combination is
\begin{align}
\mathrm{err}(\widehat{H}) = \big[\sum_{j=1}^{M} C_{jj}^{-1}\big]^{-1},
\end{align}
corresponding to weighting each regressor inversely to its error variance. The normalized optimal weights are then given by
\begin{align}
w_j = \frac{C_{jj}^{-1}}{\sum_{j'=1}^{M} C_{j' j'}^{-1}},
\end{align}
\end{lemma}

Thus, regressors with smaller error variances $C_{jj}$ receive proportionally larger weights.

Theorem~\ref{theorem:weighted-regression} provides a theoretical foundation for weighted regression combination, with Lemma~\ref{lemma:idealweights} further specifying a particular solution. However, this simplified weighting scheme relies on Assumption~\ref{assumption:zeromean-uncorr}, the stringent assumption of zero-mean and uncorrelated errors, which rarely holds in practice. To comprehensively aggregate continuous scores from multiple independent regressors, weighted combinations must account for the shift in scale in the label space. This necessitates explicit modeling of a bias term in the aggregation process to calibrate the combined output to the true label distribution.

\subsection*{StackingNet for regression combination}

\subsubsection*{Architecture}

StackingNet is an artificial neural network architecture that constructs a meta-combination from the predictions of multiple base regressors. For a query $\mathbf{x}$, the outputs of the base regressors $\{h_j\}_{j=1}^{M}$ form a continuous vector $\mathbf{h}(\mathbf{x}) = [h_1(\mathbf{x}), \ldots, h_M(\mathbf{x})]^{\mathsf{T}} \in \mathbb{R}^{M}$. StackingNet contains two sets of learnable parameters: a scalar bias term $b \in \mathbb{R}$ and per-model weights $\mathbf{w} = [w_1, \ldots, w_M]^{\mathsf{T}} \in \mathbb{R}^{M}$. As illustrated in Figure~\ref{fig:stackingnet}c, the model first applies element-wise weighting to each base prediction, then adds the global bias to the aggregated result:
\begin{align}
\widehat{H}(\mathbf{x})
= \sum_{j=1}^{M} w_j h_j(\mathbf{x}) + b
= \mathbf{w}^{\mathsf{T}} \mathbf{h}(\mathbf{x}) + b.
\label{eq:stack-arch-regression}
\end{align}
This minimalist formulation couples weights and a shift after aggregation, enabling systematic calibration of the aggregate prediction range. The parameters are initialized as $b = 0$ and $w_j = 1/M$, corresponding to the uniform-weight combination described in Theorem~\ref{theorem:regression-theory}.

\subsubsection*{Supervised learning objective}

For training the StackingNet parameters, a small amount of labeled data $\left(\mathbf{h}(\mathbf{x}), y\right) \sim \mathrm{D}_l$ are required, where $\mathbf{h}(\mathbf{x})$ denotes the vector of base predictions and $y$ the true target value. StackingNet parameters are optimized by empirical risk minimization on $\mathrm{D}_l$. The objective adopts the mean squared error (MSE) loss:
\begin{align}
\mathrm{L}_{\mathrm{regression}}(\mathbf{w}, b) = \mathbb{E}_{\left(\mathbf{h}(\mathbf{x}), y\right) \sim \mathrm{D}_l} \big[ (\widehat{H}(\mathbf{x}) - y)^{2} \big].
\label{eq:stack-loss-regression}
\end{align}
MSE objective is smooth and penalizes large deviations, facilitating stable first-order optimization. The mean absolute error (MAE) is reported solely as an more intuitive evaluation metric for interpretability.

\subsubsection*{Value range constraints}

To stabilize learning and to ensure valid parameters, StackingNet enforces non-negativity constraints element-wise:
\begin{align}
w_j \ge 0,\qquad b \ge 0.
\label{eq:constraints}
\end{align}
When the prediction values are normalized to $[0,1]$ (e.g., clipping $h_j(\mathbf{x})$ within the possible answer range, and then applying min-max normalization), a negative weight would invert a model’s contribution and encourage cancellation among regressors, whereas a negative bias could shift $\mathbf{w}^{\mathsf{T}} \mathbf{h}(\mathbf{x}) + b$ below zero. Therefore, these constraints ensure numerical stability and prevent pathological interactions. Note that $\widehat{H}(\mathbf{x})$ is not guaranteed to be no larger than $1$. Clipping is applied during inference to ensure bounded outputs. 

Note that such value range constraints are crucial for successful optimization. Supporting Information Table~\ref{tab:cfdablation} reports ablation studies for the bias term and the constraints.

\subsubsection*{Optimization}

StackingNet is computationally lightweight, containing only $(M + 1)$ learnable parameters. All labeled data can be processed in a single batch during gradient descent. The computational complexities are $\mathcal{O}(N_l)$ for bias gradients and $\mathcal{O}(N_l M)$ for weight gradients, where $N_l$ denotes the size of labeled data, which are negligible on modern hardware with small amount of labeled data. Optimization convergence is thus guaranteed in practice.

In summary, StackingNet provides a compact yet expressive meta-regressor. Unlike classic ensemble combinations that rely solely on fixed or analytically derived weights, StackingNet jointly learns per-model weights and a scalar bias. The bias term enables systematic adjustment of the aggregate prediction range, compensating for consistent over- or under-estimation in the combined base model outputs. 

\subsection*{Theoretical foundation for classification combination}

Classification differs from regression in output structure. Regressors produce continuous values, whereas classifiers output either a probability vector over classes or a hard label $h_j(\mathbf{x}) \in \{1,2,\ldots,K\}$. The latter can be represented by the one-hot indicator $h_j^k(\mathbf{x}) \in \{0,1\}$, where $h_j^k(\mathbf{x})=1$ if and only if $h_j(\mathbf{x})=k$. The following theorems establish the foundations for uniform-weight voting and weighted combinations, and the conditions under which optimal weights can be analytically obtained without annotations.

\subsubsection*{Voting}

Plurality voting selects the class receiving the largest number of votes from the base classifiers:
\begin{align}
\hat{h}_{\text{plurality}}(\mathbf{x}) := \argmax_{k \in \{1, 2, \ldots, K\}} \sum_{j=1}^M h_j^k(\mathbf{x}),
\end{align}
with ties broken uniformly at random.

\noindent
Majority voting extends plurality voting by introducing a rejection option, producing a class label only if it receives more than half of the votes:
\begin{align}
\hat{h}_{\text{majority}}(\mathbf{x}) :=
\begin{cases}
k, & \text{if } \sum_{j=1}^M h_j^k(\mathbf{x}) > \frac{M}{2} \\
\text{rejection}, & \text{otherwise}
\end{cases}.
\end{align}

For binary classification ($K=2$), majority and plurality voting coincide. The following theorem shows that the voting results improve with the number of competent classifiers.

\begin{theorem}[Monotonicity and convergence of voting with increasing number of classifiers \cite{Lam1997}] \label{theorem:votingconvergence}
Under Assumptions~\ref{assumption:iid} and~\ref{assumption:independence}, if all base classifiers have identical correct prediction probability $p>0.5$ for each test sample, then $P_{\mathrm{voting}}(M,p)$ is monotonically increasing in $M$, and
\begin{align}
\lim_{M \to +\infty} P_{\text{voting}} (M, p) = 1.
\end{align}
\end{theorem}

\subsubsection*{Weighted combination}

Because base classifiers typically have unequal performance, weighted voting assigns larger weights to more reliable classifiers.

\begin{theorem}[Optimal weights for weighted classification combination \cite{Shapley1984, Lam1997}] \label{theorem:weighted-voting}
Let $p_j$ denote the (unknown) correct prediction probability of base classifier $h_j$ on the test data distribution. Under Assumptions~\ref{assumption:iid} and ~\ref{assumption:independence} and for $K=2$, the optimal weights satisfy
\begin{align}
w_j \propto \log \frac{p_j}{1-p_j}. \label{eq:optimal-weights}
\end{align}
\end{theorem}

Thus, the weights are proportional to the log-odds of accuracy: more accurate classifiers receive larger weights. Note that formula (\ref{eq:optimal-weights}) does not, by itself, account for class priors. It also does not consider the actual test samples, assuming a general correct prediction probability on test data distribution.

In practice, estimating the proportionality constant in Theorem~\ref{theorem:weighted-voting} requires knowledge of classifier reliability. When labeled data are available, reliability can be estimated from empirical sensitivity and specificity, for example via the balanced accuracy $\pi_j$, considering class priors. When labels are unavailable, spectral methods can approximate these quantities, as described next.

\subsubsection*{Unsupervised reliability estimation}

Unsupervised estimation of classifier reliability can be achieved by spectral decomposition of prediction covariance. For brevity, we focus on the binary case where $h_j(\mathbf{x}) \in \{-1,1\}$ and equivalently $h_j^k(\mathbf{x}) \in \{0,1\}$. Let $Q$ denote the $M\times M$ population covariance matrix of classifier outputs on test samples.

\begin{lemma}[Covariance between classifiers \cite{Parisi2014}] \label{lemma:covariance}
Under Assumptions~\ref{assumption:iid} and \ref{assumption:independence}, the entries of $Q$ satisfy
\begin{align}
q_{j j'} =     \begin{cases}
      1 - \mu_{j}^{2}, &  j=j' \\
      (2 \pi_j - 1)(2 \pi_{j'} - 1)(1 - b^{2}), & j \neq j'
    \end{cases}, \label{eq:q-binary}
\end{align}
where $\mu_j = \mathbb{E}[h_j(\mathbf{x})]$ denotes the population mean, $b = \mathbb{P}[Y = 1] - \mathbb{P}[Y = -1]$ denotes the difference in class priors, i.e., class imbalance, and $\pi = (\mathbb{P}[h(\mathbf{x})=Y | Y=1] + \mathbb{P}[h(\mathbf{x})=Y | Y=-1]) / 2$ denotes the balanced classification accuracy of classifier $h_j$, respectively.
\end{lemma}

Importantly, the off-diagonal entries of $Q$ form a rank-one matrix that can be decomposed as $R= \lambda \mathbf{v}\mathbf{v}^{\mathsf{T}}$, where the unit-norm eigenvector $\mathbf{v} \in \mathbb{R}^M$ and the eigenvalue $\lambda = (1 - b^2) \sum_{j=1}^M (2\pi_j - 1)^2$. This leads to the deduction of the following theorem.

\begin{theorem}[Spectral decomposition for classifier reliability estimation \cite{Parisi2014}] \label{theorem:binary-sml}
Under Assumptions~\ref{assumption:iid} and~\ref{assumption:independence}, up to a sign ambiguity, the entries of the principal eigenvector $\mathbf{v}$ are proportional to the balanced classification accuracies of the base classifiers on the test data
\begin{align}
v_j \propto (2\pi_j - 1), \label{eq:smlproportion}
\end{align}
\end{theorem}

If, additionally, Assumption~\ref{assumption:majorcompetence} can be fulfilled where a majority of base classifiers outperform random guessing, then the entries of $\mathbf{v}$ can be used to form the weighted combination, with weights $v_j$ proportional to $\pi_{j}$ of the base classifiers.

Theorems~\ref{theorem:weighted-voting} and~\ref{theorem:binary-sml} jointly establish the theoretical foundation for weighted classification combination. Spectral estimates can provide unsupervised reliability initialization, whereas supervised reliability can also be derived directly from labeled data. To our knowledge, no existing approach has unified both supervised and unsupervised reliability estimation within an identical data-driven learning framework.

\subsection*{StackingNet for classification combination}

\subsubsection*{Architecture}

StackingNet for classification is a lightweight neural architecture that constructs a meta-combination from the class predictions of multiple base classifiers. For an input sample $\mathbf{x}$, the outputs of the base classifiers $\{h_j\}_{j=1}^{M}$ are one-hot vectors $h_j(\mathbf{x}) \in \{0,1\}^K$. These vectors are stacked into a prediction matrix $\mathbf{h}(\mathbf{x}) = [h_1(\mathbf{x}), \ldots, h_M(\mathbf{x})] \in \{0,1\}^{K \times M}$, where each column corresponds to a base classifier’s prediction. StackingNet employs a single set of learnable parameters $\mathbf{w} = [w_1, \ldots, w_M]^{\mathsf{T}} \in \mathbb{R}^M$, with $w_j$ representing the reliability of the $j$-th classifier. As illustrated in Figure~\ref{fig:stackingnet}c, the combined class score vector is obtained as
\begin{align}
\widehat{\mathbf{H}}(\mathbf{x})
= \sum_{j=1}^{M} w_j h_j(\mathbf{x})
= \mathbf{h}(\mathbf{x}) \mathbf{w}.
\label{eq:stack-arch-classification}
\end{align}
This formulation directly links each classifier’s contribution to its reliability weight, ensuring that more accurate classifiers exert stronger influence in the final decision. Unlike in regression, bias terms are unnecessary because classification depends on relative rather than absolute score calibration. The weights are initialized as $w_j = 1/M$, corresponding to uniform-weight voting.

\subsubsection*{Supervised learning objective}

The weight vector $\mathbf{w}$ is learned via empirical risk minimization on labeled data $(\mathbf{h}(\mathbf{x}), y) \sim \mathrm{D}_l$, where $y \in \{1, \ldots, K\}$ is the ground-truth label and $\mathbf{y} \in \{0,1\}^K$ its one-hot representation. The supervised learning objective adopts the classic cross-entropy loss formulation:
\begin{align}
\mathrm{L}_{\mathrm{sup}}(\mathbf{w})
= \mathbb{E}_{(\mathbf{h}(\mathbf{x}), y) \sim \mathrm{D}_l}
\left[
-\sum_{k=1}^{K} y^k
\log
\frac{
\exp\left(\widehat{\mathbf{H}}^k(\mathbf{x})\right)
}{
\sum_{k'=1}^{K} \exp\left(\widehat{\mathbf{H}}^{k'}(\mathbf{x})\right)
}
\right].
\label{eq:stack-loss-sup}
\end{align}
This objective encourages correct predictions via combination while producing smooth gradients for stable optimization. When labeled data are class-imbalanced, weighted cross-entropy loss adopts inverse of class frequency in $\mathrm{D}_l$.

\subsubsection*{Unsupervised learning objective}
Different from regression, the weights $\mathbf{w}$ can also be learned on unlabeled data $\mathbf{h}(\mathbf{x}) \sim \mathrm{D}$, rooted in insights from Theorems~\ref{theorem:weighted-voting} and~\ref{theorem:binary-sml} to align base classifiers with the aggregated prediction.

StackingNet generates a consensus pseudo-label $\hat{\mathbf{y}} \in \{0,1\}^K$ from its combined score $\widehat{\mathbf{H}}(\mathbf{x})$:
\begin{align}
\hat{y}^k = \begin{cases} 1, & \text{if } k = \argmax_{k'} \widehat{\mathbf{H}}^{k'}(\mathbf{x})
 \\ 0, & \text{otherwise} \end{cases}.
\label{eq:consensus-label}
\end{align}

The unsupervised loss minimizes the disagreement between each base classifier and the aggregated consensus:
\begin{align}
\mathrm{L}_{\mathrm{unsup}}(\mathbf{w}) = \mathbb{E}_{\mathbf{h}(\mathbf{x}) \sim \mathrm{D}} \bigg[ \sum_{j=1}^M w_j \cdot \mathbb{I}\big[ h_j(\mathbf{x}) \neq \hat{\mathbf{y}} \big] \bigg].
\label{eq:stack-loss-unsup}
\end{align}
where $\mathbb{I}[\cdot]$ is the indicator function. This objective gradually increases the influence of more reliable classifiers as learning progresses.

\subsubsection*{Value range constraints}
To ensure a convex combination, StackingNet enforces two constraints on the weight vector. First, the weights must remain non-negative:
\begin{align}
w_j \ge 0, \qquad j = 1, \ldots, M.
\label{eq:class-constraints}
\end{align}

The second ensures the weight vector sums up to 1, preventing unbalanced scaling and maintaining consistency with reliability interpretation. The regularization term is
\begin{align}
\mathrm{L}_{\mathrm{reg}}(\mathbf{w}) = \big(1 - \sum_{j=1}^M w_j \big)^2
\label{eq:class-regularization}
\end{align}

Note that to accelerate convergence and to ensure stability in convex optimization, the learnable weights of StackingNet can be initialized in two ways other than uniform-weight initialization. When labeled data are available, BCA scores can be directly calculated on labeled data and be used after normalized as initialization. When only unlabeled data are available, the normalized BCA scores of voting aggregation can be used instead.

\subsubsection*{Optimization}

The aforementioned learning objectives can be integrated. When both labeled and unlabeled data are available, the overall objective is
\begin{align}
\mathrm{L}_{\mathrm{classification}}(\mathbf{w})
= \mathrm{L}_{\mathrm{sup}}(\mathbf{w})
+ \lambda_1 \cdot \mathrm{L}_{\mathrm{unsup}}(\mathbf{w}) + \lambda_2 \cdot \mathrm{L}_{\mathrm{reg}}(\mathbf{w}),
\label{eq:stack-loss-semi}
\end{align}
where $\lambda_1$ and $\lambda_2$ are hyperparameters. The coefficient $\lambda_1$ balances the supervised and unsupervised losses on comparable scales, whereas $\lambda_2$ controls the regularization strength to ensure smooth weight normalization. As Supporting Information Figure~\ref{fig:weight-sensitivity} shows, incorporating these additional objectives proves beneficial, and in some cases essential.

In summary, StackingNet for classification adopts a compact architecture yet expressive learning mechanisms for classification combination. It unifies supervised and unsupervised combination within a single optimization framework, enabling robust meta-combination under both labeled and unlabeled regimes. It maintains explainability that the weight vector it learned can be further utilized for ranking and pruning.

\subsection*{Method comparison}

To evaluate the performance of the proposed StackingNet framework, we compare it to linear classifier/regressor of classic machine learning as of traditional stacked generalization~\cite{Wolpert1992}, instead of neural network models. Note that these approaches are limited to supervised learning.

For classification, StackingNet was further compared against a wide range of unsupervised combination methods:
\begin{enumerate}
\item Voting, which employs plurality voting, where the class receiving the highest number of votes is selected, even if it does not achieve an absolute majority. Ties are resolved by random selection.
\item Worker Agreement with Aggregate (WAwA), which first computes the majority vote label and then estimates each base classifier’s reliability as the proportion of its predictions that agree with this consensus. These reliability scores are subsequently used to perform a weighted majority vote. When ground-truth annotations are available, classifier reliabilities can be directly estimated from labeled data.
\item Dawid-Skene~\cite{Dawid1979}, which is a probabilistic model that jointly estimates the true labels and classifier reliability by learning a confusion matrix for each classifier. It is optimized via the expectation-maximization (EM) algorithm to iteratively infer latent true labels and update classifier-specific error rates.
\item Generative model of Labels, Abilities, and Difficulties (GLAD)~\cite{Whitehill2009}, which uses a probabilistic model to simultaneously model the three elements, optimized with EM.
\item Multi-Annotator Competence Estimation (MACE)~\cite{Hovy2013}, which is a probabilistic method that accounts for spamming behavior by modeling label distributions for unreliable annotators, optimized using EM.
\item Matrix-Mean-Subsequence-Reduced (M-MSR)~\cite{Ma2020}, which estimates classifier reliability by filtering out extreme values in iterative alternating updates, handling adversarial classifiers that deviate arbitrarily from the Dawid-Skene model.
\item Karger, Oh, and Shah (KOS)~\cite{Karger2011KOS}, which is an iterative algorithm that calculates the log-likelihood of the sample being positive while modeling the base model's reliability. It is only applicable to binary classification.
\item ZenCrowd~\cite{Demartini2012}, which uses the probabilistic graphical model to create micro-tasks for uncertain cases and applies EM-based inference over factor graphs to estimate true links and classifier reliability, optimized with EM.
\item Participant-Mine voting (PM)~\cite{Aydin2014, Li2014}, which iteratively optimizes for truths and per-classifier reliability by minimizing weighted disagreement between observations and inferred truths.
\item Label Aggregation (LA)~\cite{Yang2024a} uses a dynamic Bayesian network to estimate classifier qualities and true labels by traversing all labels twice.
\item Label-Aware Autoencoders (LAA)~\cite{Yin2017} uses a neural network model, integrating an encoder and a decoder to infer true labels through minimizing a reconstruction loss, optimized by backpropagation.
\item Enhanced Bayesian Classifier Combination (EBCC)~\cite{Li2019} iteratively optimizes a variational inference framework that models each true label as a mixture of latent subtypes, where classifier reliability varies by subtype.
\item Spectral Meta-Learner (SML)~\cite{Parisi2014} uses spectral decomposition for unsupervised reliability estimation. In multi-class case, one-vs-rest extension was used to approximate the combination weights~\cite{Li2025SML-OVR}.
\end{enumerate}

For regression, we compared StackingNet with uniform-weight averaging, which, to the best of our knowledge, is the only reliable alternative to the proposed framework. In practice, unsupervised regression combination remains a fundamentally difficult problem.

\subsection*{Data and experiments}

We evaluated StackingNet across three groups of datasets:
\begin{enumerate}
\item The first group includes four sub-datasets, ICLR2025, ICLR2024, NeurIPS2024 and NeurIPS2023, which consist of texts converted from research paper PDFs as inputs. For each paper, we queried six proprietary API-based LLMs, and also collected public reviews from at least three anonymous human reviewer. The scalar regression scores on the final overall rating of the paper were used.
\item The second group consists of thirteen sub-datasets from the Chicago Face Database (CFD)~\cite{Ma2015}, where facial images are used as inputs. As the original release from CFD forbids querying the facial images to open-source LLMs, we locally deployed ten VLMs to generate continuous ratings for attributes on dimensions of afraid, angry, babyfaced, feminine, masculine, sad, threatening, unusual, attractive, disgusted, happy, surprised, and trustworthy. Ground-truth rating scores are the consensus average over many human raters per face. In the norming study, 1{,}087 raters each judged a random subset of ten to fifteen faces, so that each face was scored by roughly seventy raters on average~\cite{Ma2015}. The VLMs range from 2B to 7B parameters.
\item The third group includes eight sub-datasets from the Holistic Evaluation of Language Models (HELM) benchmark~\cite{Liang2023}, covering binary and multi-class classification tasks on general question answering, reasoning, multiple choice question, etc. Sub-datasets include BoolQ, EntityMatching, IMDB, LegalSupport, LSAT, MMLU, RAFT, and CivilComments. We used query outputs from ten proprietary API-based LLMs already included in the HELM release.
\end{enumerate}

Table~\ref{tab:datastatistics} provides an overview of dataset statistics. Supporting Information Table~\ref{tab:llmstatistics} provides complete information on the queried LLMs/VLMs. The exact prompt text that were used to generate the responses are detailed in Supporting Information Materials.

\begin{table}[htpb]  \centering
\caption{Dataset statistics.}  \label{tab:datastatistics}
\resizebox{\textwidth}{!}{\begin{tabular}{lllll}
\toprule
Dataset Name & \# Samples & Task Type & Range / Class & Queried LLMs/VLMs \\
\midrule
ICLR2025 & 1,000 & \multirow{4}{*}{\shortstack[l]{research paper rating}} & \multirow{4}{*}{[1, 10]} & \multirow{4}{*}{\shortstack[l]{[API-based LLMs]\\DeepSeek-R1, Doubao,\\Gemini-2, GPT-4o,\\GPT-5, Qwen-Turbo}} \\
ICLR2024 & 1,000 & & & \\
NeurIPS2024 & 1,000 & & & \\
NeurIPS2023 & 1,000 & & & \\
\midrule
\multirow{4}{*}{\shortstack[l]{Chicago\\Face\\Database\\\\(CFD)}} & \multirow{4}{*}{\shortstack{1,441}} & \multirow{4}{*}{\shortstack[l]{attribute rating of \\human facial images \\in 13 dimensions}} & \multirow{4}{*}{\shortstack{[1, 7]}} & \multirow{4}{*}{\shortstack[l]{[local VLMs]\\BLIP, DeepSeek-VL, H2OVL,\\InternVL-2, LLaVA, Molmo,\\Paligemma, Phi-3.5, SAIL-VL, SmolVLM}} \\
&&&&\\
&&&&\\
&&&&\\
\midrule
BoolQ & 5,000 & question answering & yes / no & \multirow{8}{*}{\shortstack[l]{[API-based LLMs]\\AnthropicLM-v4, Command,\\Falcon, GPT-3.5-Turbo,\\Jurassic-2, Llama-2,\\MPT, Palmyra-X,\\RedPajama, TNLG-v2,}} \\
CivilComments & 45,000 & question answering & yes / no & \\
EntityMatching & 1,400 & question answering & yes / no & \\
IMDB & 1,000 & sentiment choice & positive / negative  & \\
LegalSupport & 1,000 & question answering & binary choice & \\
LSAT & 461 & multiple choice & 5 choices & \\
MMLU & 1,641 & multiple choice & 4 choices & \\
RAFT & 1,348 & question answering & binary choice & \\
\midrule
\multicolumn{5}{c}{Total Number of Queries: 24,000 + 187,330 + 568,500} \\
\bottomrule
\end{tabular}}
\begin{tablenotes}
\footnotesize
\item \textit{LLMs/VLMs are ordered alphabetically.}
\end{tablenotes}
\end{table}

\clearpage

\section*{Acknowledgements}
This work was supported by Brain Science and Brain-Like Intelligence Technology-National Science and Technology Major Project 2021ZD0201300, and National Natural Science Foundation of China 62525305. 

The authors thank Haoqing Jiang for helpful discussions on large language models.

\section*{Author Contributions}
S.L.\ and C.L.\ contributed equally to this work. S.L.\ conceived the study and wrote the manuscript. S.L.\ and C.L.\ designed the methods, conducted the experiments, and analyzed the results. D.W., Z.Z., and L.D.\ supervised and funded the study. All authors reviewed the manuscript.

\section*{Conflict of Interest}
The authors declare no conflict of interest.

\section*{Data Availability Statement}
\begin{enumerate}
\item All post-processed predictions from LLMs/VLMs are provided as \texttt{.csv} files and publicly available on GitHub (\url{https://github.com/sylyoung/TestEnsemble}).
\item The locally deployed VLMs used can be accessed and downloaded via Hugging Face (\url{https://huggingface.co/models}) by searching for the complete model version IDs.
\item The raw PDF manuscripts and associated anonymous human reviews from ICLR and NeurIPS are available through OpenReview (https://openreview.net/). The list of paper IDs used is available on GitHub (\url{https://github.com/sylyoung/TestEnsemble}).
\item Access to the original facial image data from the Chicago Face Database (CFD)~\cite{Ma2015} requires permission and can be requested at \url{https://www.chicagofaces.org/}.
\item The raw text prompt entries for the eight datasets from HELM benchmark~\cite{Liang2023} were obtained from HELM Release v0.4.0 (2023-11-17), available at \url{https://crfm.stanford.edu/helm/classic/latest/}.
\item All codes that were used in this paper are publicly available on GitHub (\url{https://github.com/sylyoung/TestEnsemble}).
\end{enumerate}

\clearpage

\bibliographystyle{advsci}
\bibliography{stackingAS}

\clearpage

\renewcommand{\thefigure}{S\arabic{figure}}
\renewcommand{\figurename}{Supporting Information Figure}
\setcounter{figure}{0}

\renewcommand{\thetable}{S\arabic{table}}
\renewcommand{\tablename}{Supporting Information Table}
\setcounter{table}{0}

\section*{Supporting Information} \label{sect:prompts}

\begin{figure}[htpb]\centering
\includegraphics[width=\textwidth]{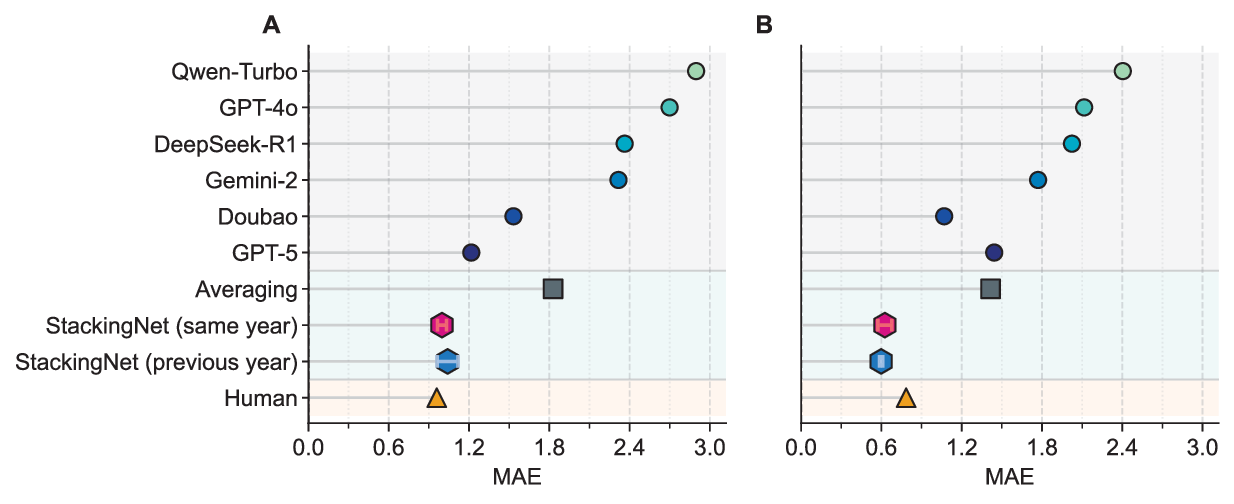}
\caption{{\bfseries Research paper rating error by individual human reviewers, individual LLMs, and collective inference of multiple LLMs.}
{\bfseries a-b,} Mean absolute error (MAE, lower is better) across two datasets, ICLR2025 and NeurIPS2024. StackingNet was trained in a few-shot setting using 1\% of labeled examples (10 papers with ground-truth scores from multiple human reviewers) drawn either from the same year or from the previous year.}
\label{fig:paperreview-supp}
\end{figure}

\clearpage

\begin{figure}[htpb]
\centering
\includegraphics[width=\textwidth]{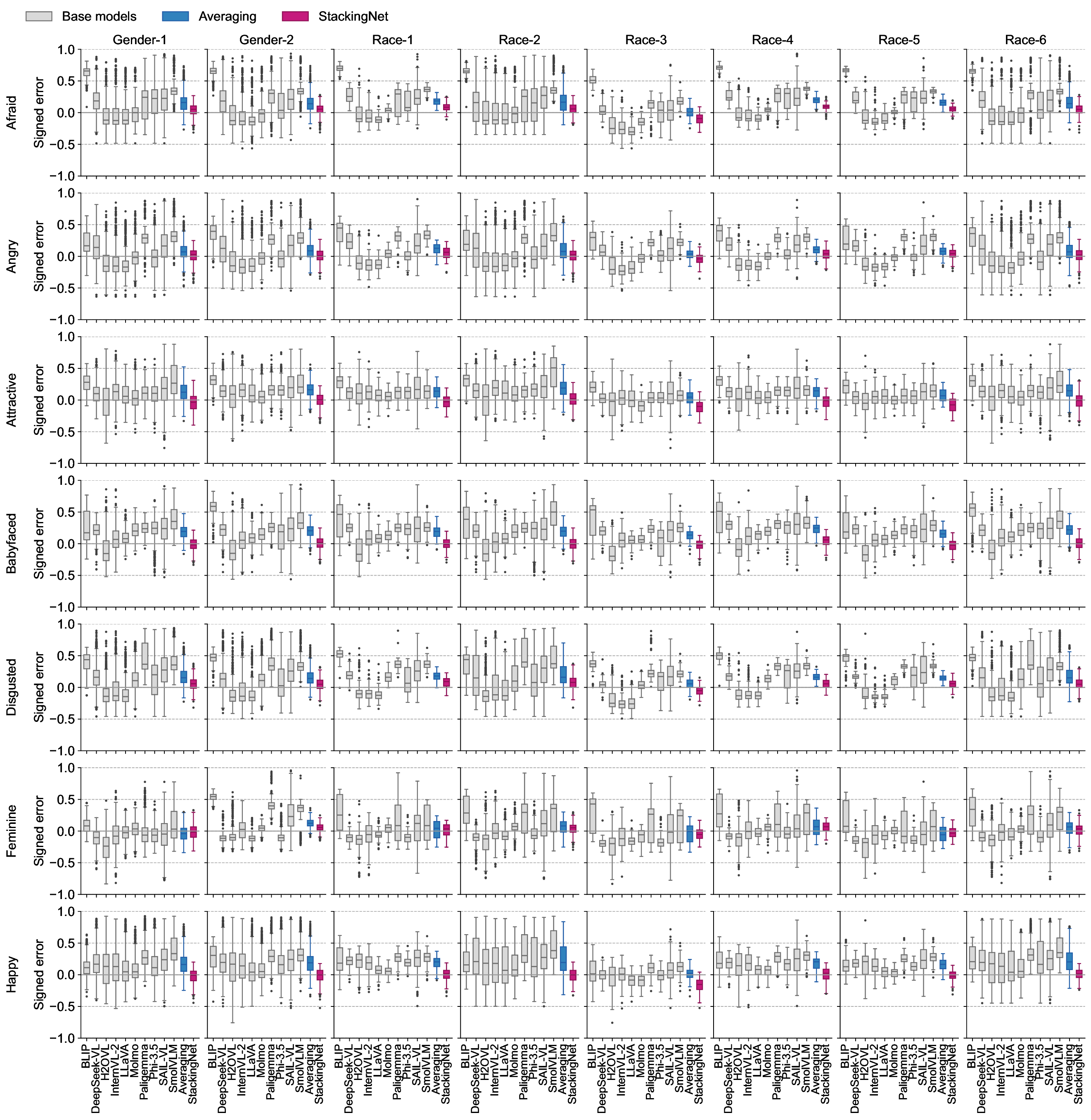}
\caption*{(\textit{Supporting Information Figure 2 continued on next page})}
\end{figure}

\begin{figure}[htpb]
\centering
\includegraphics[width=\textwidth]{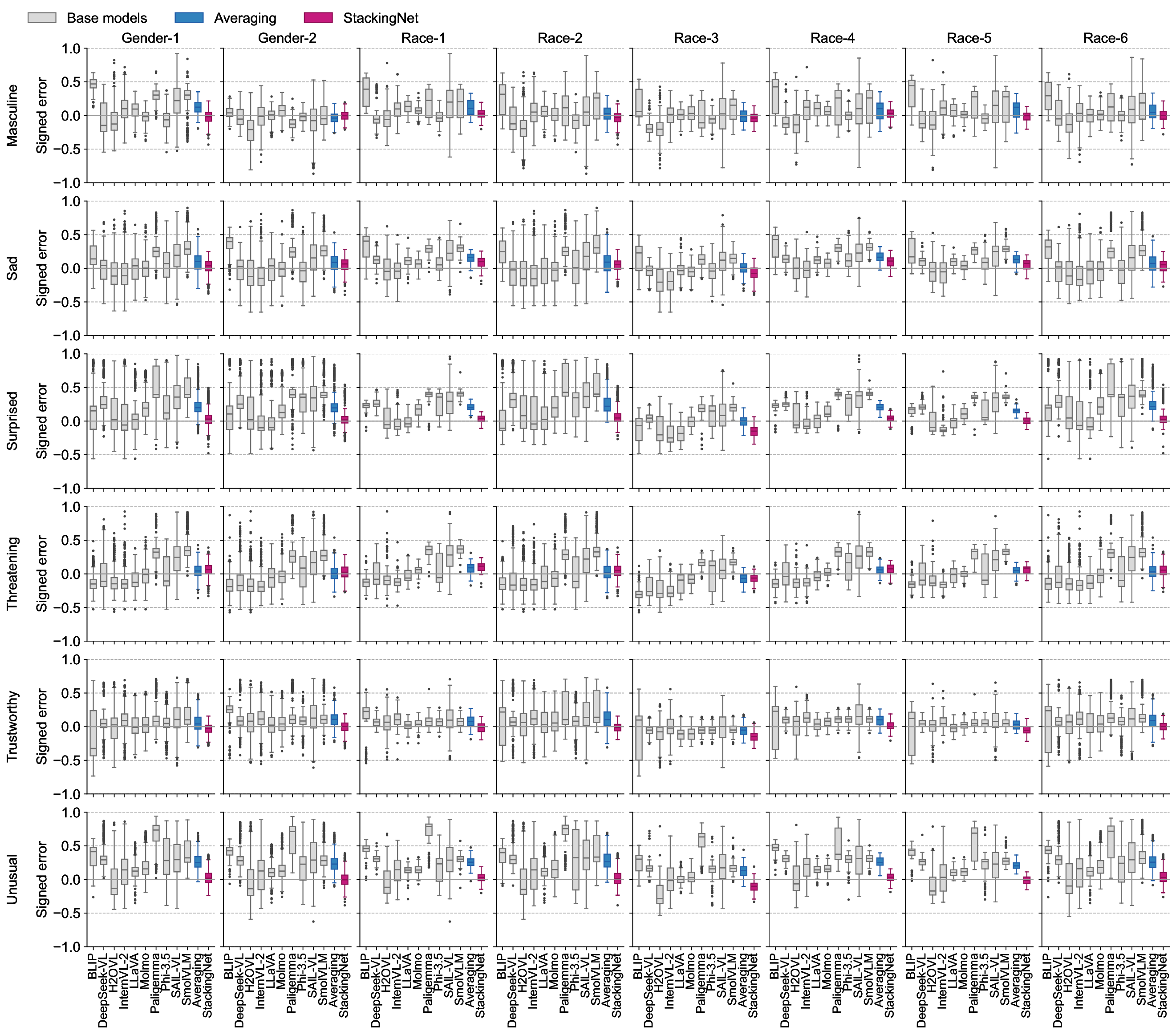}
\caption{{\bfseries Facial image attribute ratings by VLMs and StackingNet combination on the Chicago Face Database.}
Predicted ratings are shown across gender and race/ethnicity groupings.}
\label{fig:cfd-supp}
\end{figure}

\clearpage

\begin{figure}[htpb]\centering
\includegraphics[width=\textwidth]{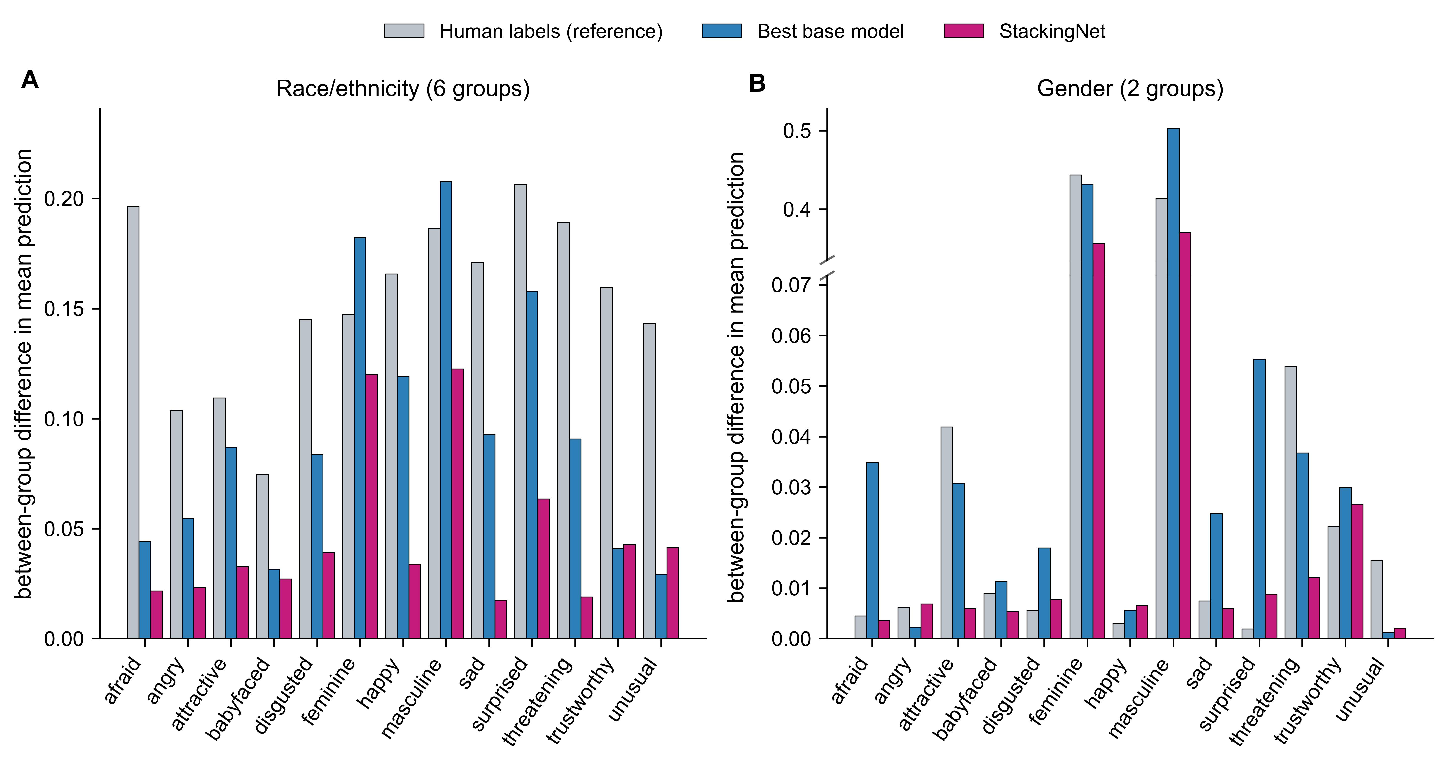}
\caption{{\bfseries Demographic-parity view of the Chicago Face Database facial-attribute task.}
{\bfseries a}, the six race/ethnicity groups. {\bfseries b}, the two gender groups.
Between-group difference in the mean predicted score, the largest minus the smallest per-group mean prediction~\cite{Chzhen2020}, for the human reference labels, the best single base model, and StackingNet, in the normalized $[0,1]$ label space. The plotted values are absolute predicted scores rather than errors relative to the human labels, although StackingNet is still trained on the 1\% human-annotated labels.}
\label{fig:fairness-dp}
\end{figure}

\clearpage

\begin{figure}[htpb]\centering
\includegraphics[width=\textwidth]{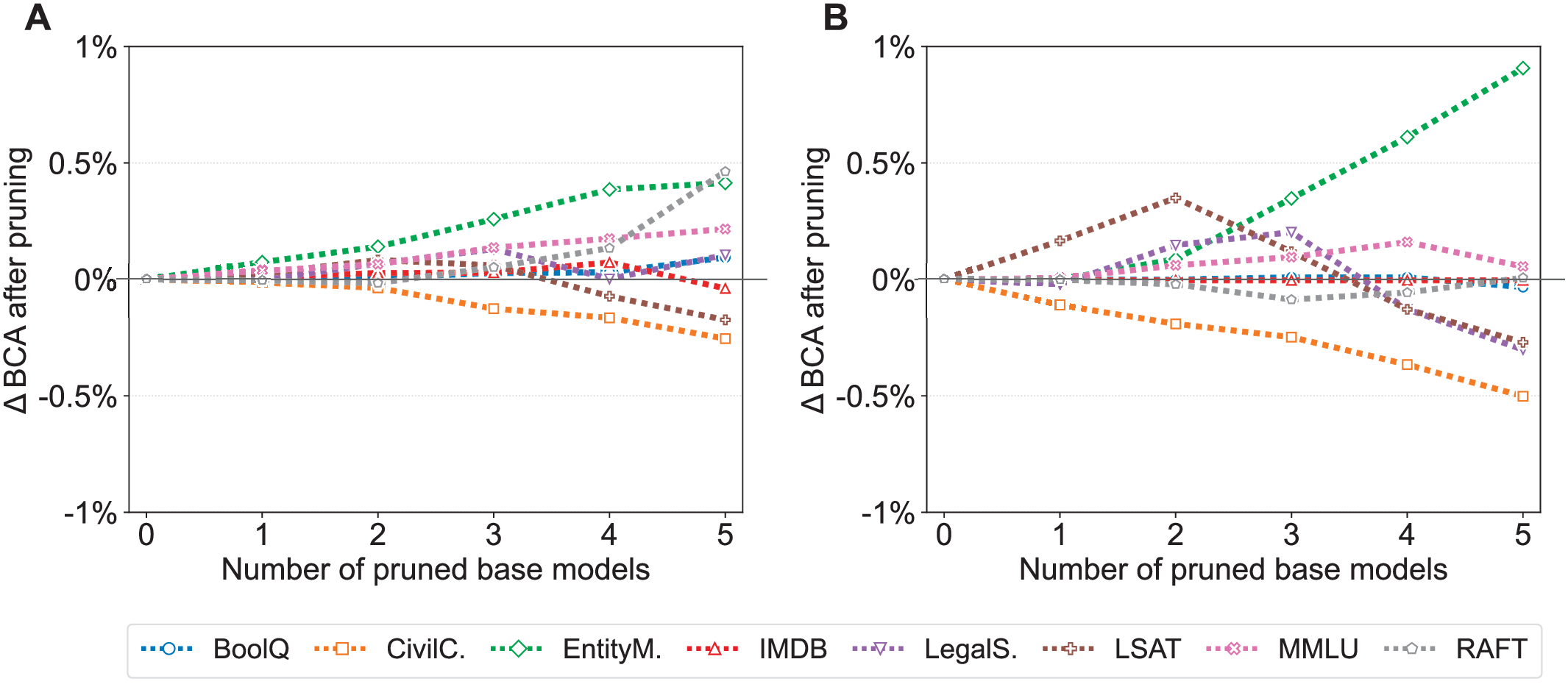}
\caption{{\bfseries Performance of StackingNet on classification tasks for adversary pruning.}
Relative changes in BCA after sequentially pruning the lowest-weighted base models are shown.
{\bfseries a}, Performance under random-injection attacks. Five randomly predicting base models are injected into the combination.
{\bfseries b}, Performance under label-flipping attacks. Five low-performing base models in the original combination are flipped to predict a different class.
Results are averaged over one hundred runs with different random seeds.}
\label{fig:attackprune-supp}
\end{figure}

\clearpage

\begin{figure}[htpb]\centering
\includegraphics[width=\textwidth]{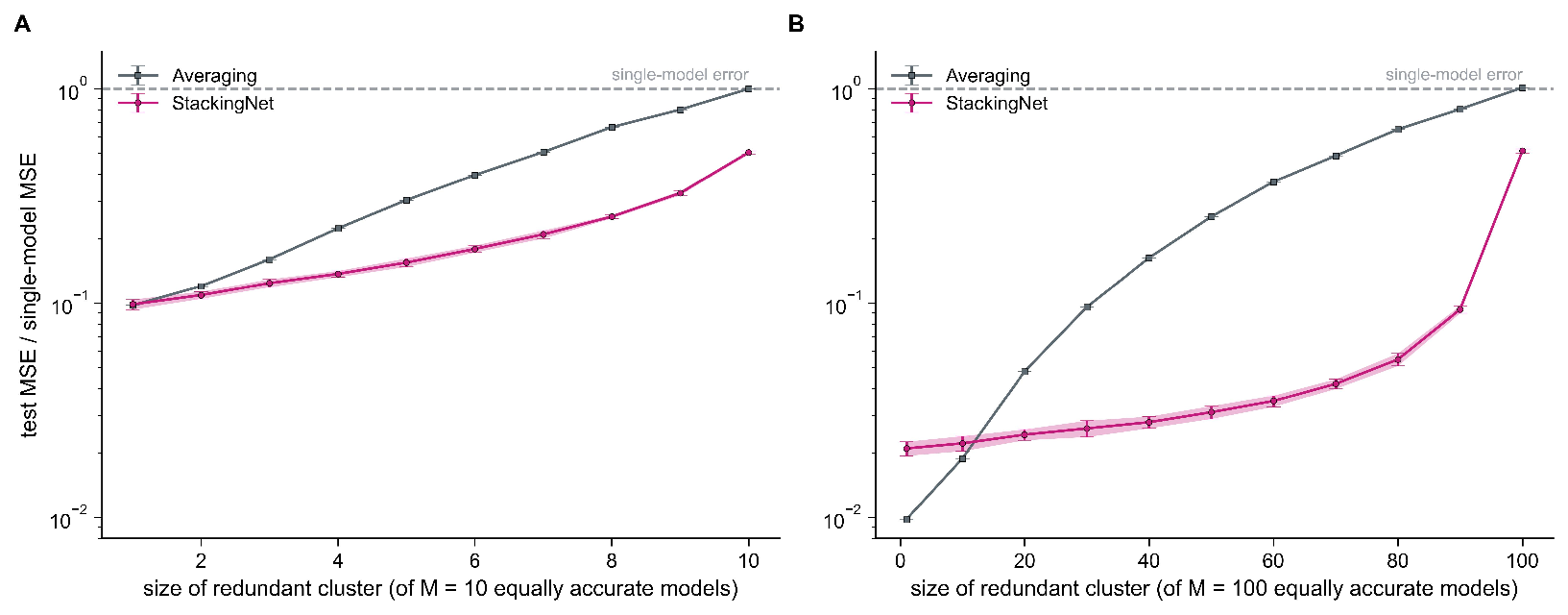}
\caption{{\bfseries Test error of averaging and StackingNet as a redundant cluster of base models grows.} Synthetic data: a pool of $M$ regressors of equal error variance, of which a cluster of $c$ models is made mutually redundant by sharing one common error term while the remaining $M-c$ keep independent errors. {\bfseries a}, $M=10$. {\bfseries b}, $M=100$. The vertical axis is the test mean-squared error relative to the single-model MSE on a logarithmic scale, so the dashed line at one marks the error of any single base model. Each pool draws 10,000 examples, of which 1\% are labeled to fit StackingNet. Markers and shaded bands are the mean and standard deviation over 30 repetitions that each resample the labeled subset and reinitialize StackingNet.}
\label{fig:degradation}
\end{figure}

\clearpage

\begin{figure}[htpb]\centering
\includegraphics[width=\textwidth]{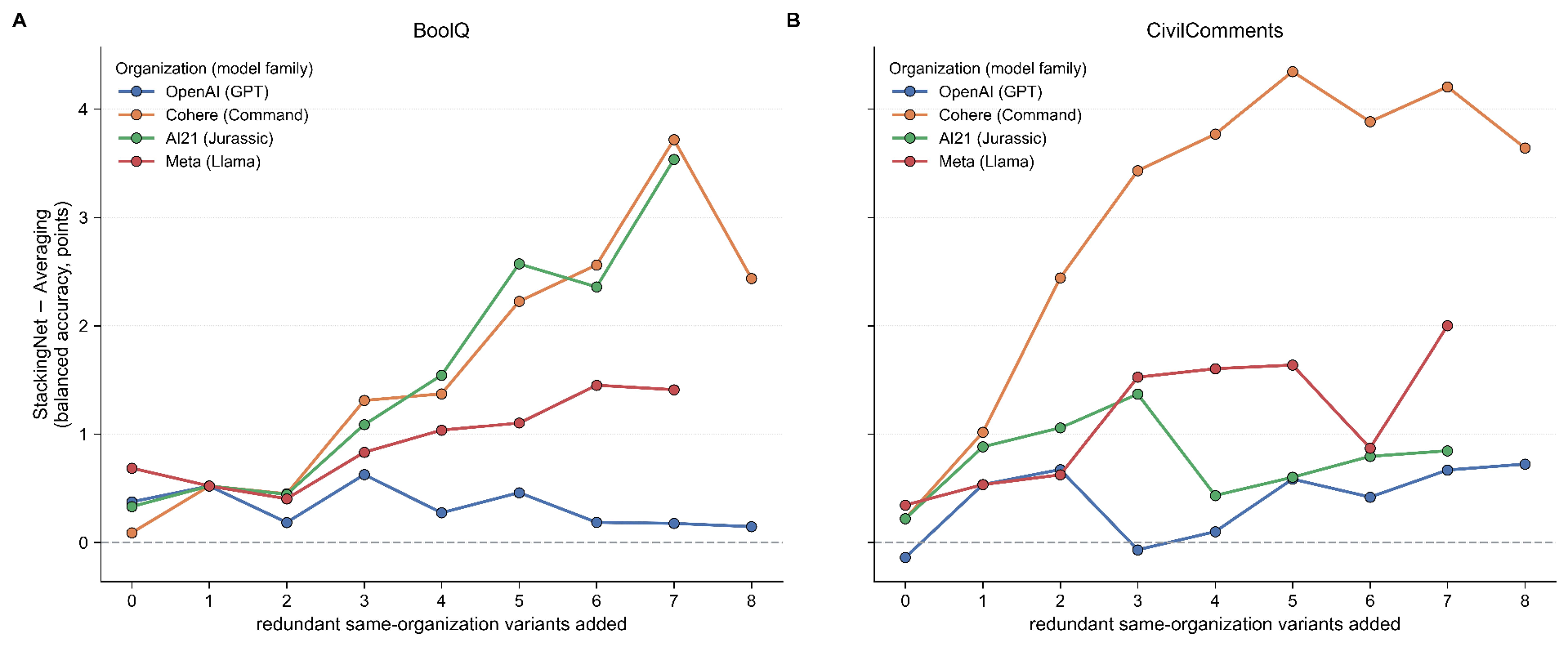}
\caption{{\bfseries Effect of adding redundant model variants from a single organization to a diverse pool.} {\bfseries a}, BoolQ. {\bfseries b}, CivilComments. Starting from a fixed pool of five strong models, one from each of five organizations, an increasing number of further variants from one organization is added, that is, models from the same developer that share architecture, training data, and tuning recipe and therefore make correlated errors. Each curve is one such organization. The vertical axis is StackingNet's balanced accuracy minus that of unweighted averaging, in percentage points, against the number of redundant variants added. Both methods use a 5\% labeled split, averaged over five seeds, and the dashed line marks equality.}
\label{fig:helm-robust}
\end{figure}

\clearpage

\begin{figure}[htpb]\centering
\includegraphics[width=\textwidth]{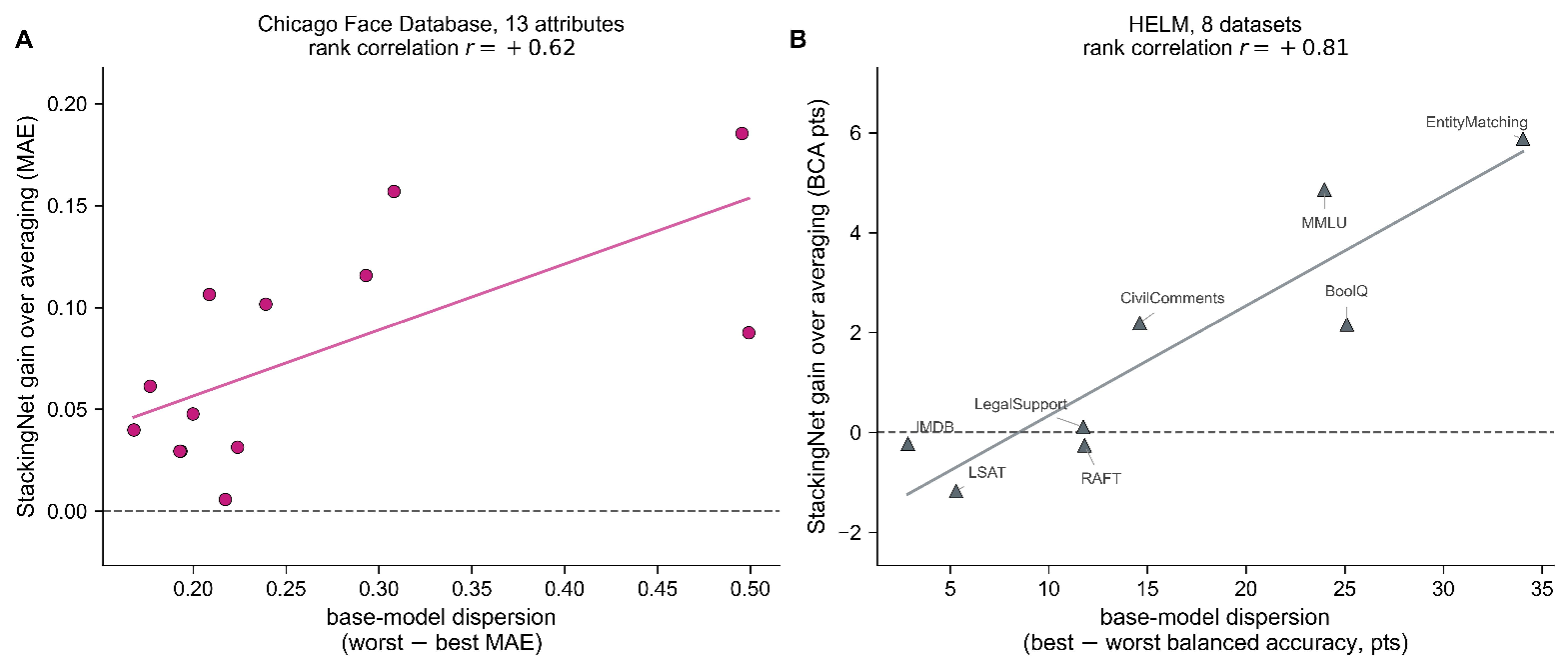}
\caption{{\bfseries StackingNet's gain over unweighted averaging versus base-model dispersion.}
{\bfseries a}, the thirteen Chicago Face Database facial-attribute regression tasks, in MAE. {\bfseries b}, the eight HELM classification datasets, each labeled by name, in balanced-accuracy points.
The horizontal axis is base-model dispersion, the spread between the best and worst base model in the pool. The vertical axis is StackingNet's gain over unweighted averaging of the base models, so a value above the dashed zero line indicates that StackingNet outperforms averaging. In each panel the line is a least-squares fit and the rank correlation coefficient is shown.}
\label{fig:dispersiongain}
\end{figure}

\clearpage

\begin{figure}[htpb]\centering
\includegraphics[width=\textwidth]{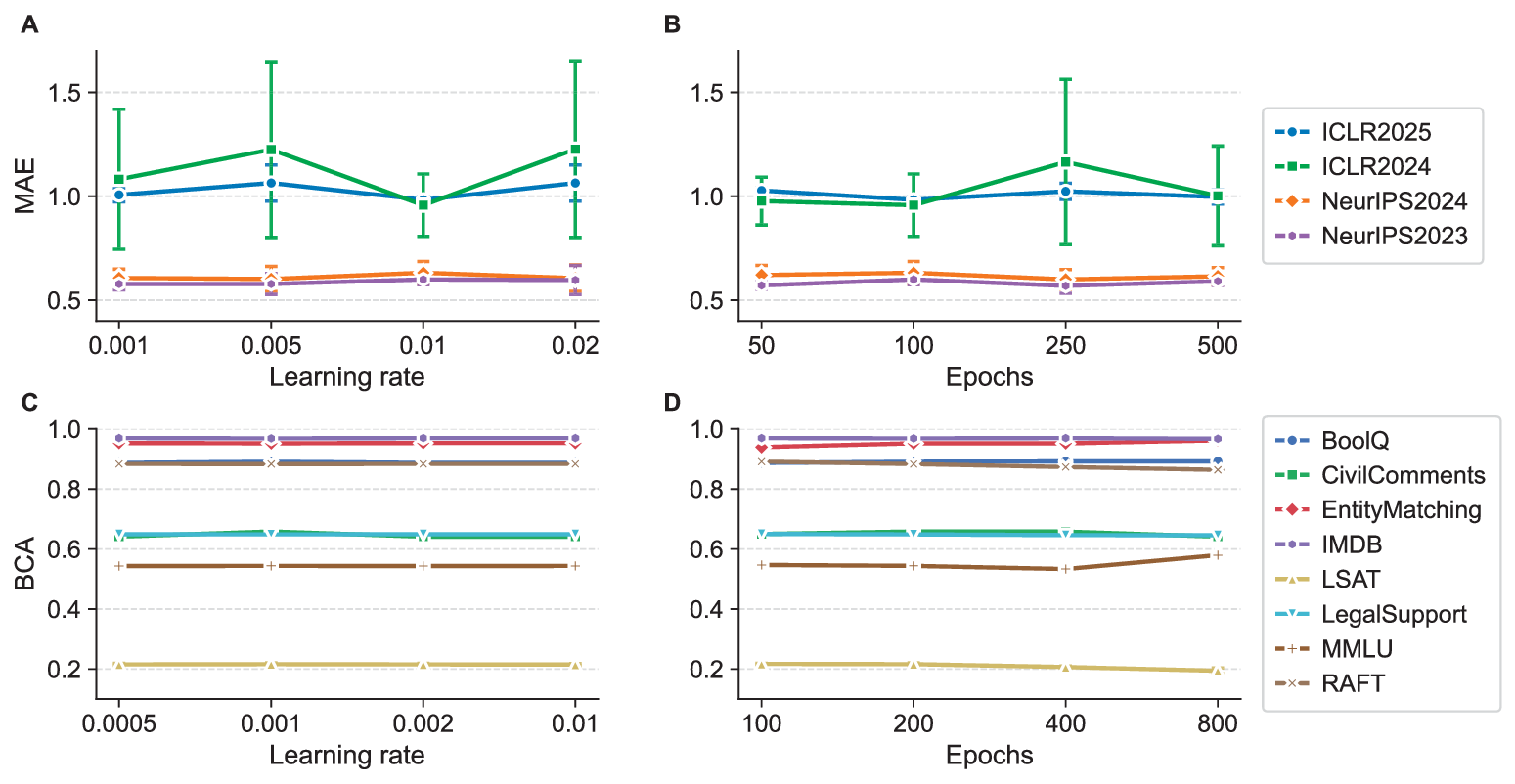}
\caption{{\bfseries Sensitivity of StackingNet to different learning rates and training epochs.}
{\bfseries a-b}, MAE of StackingNet on regression datasets w.r.t. different learning rate and training epochs.
{\bfseries c-d}, BCA of StackingNet on classification datasets w.r.t. different learning rate and training epochs.}
\label{fig:sensitivity}
\end{figure}

\clearpage

\begin{figure}[htpb]\centering
\includegraphics[width=\textwidth]{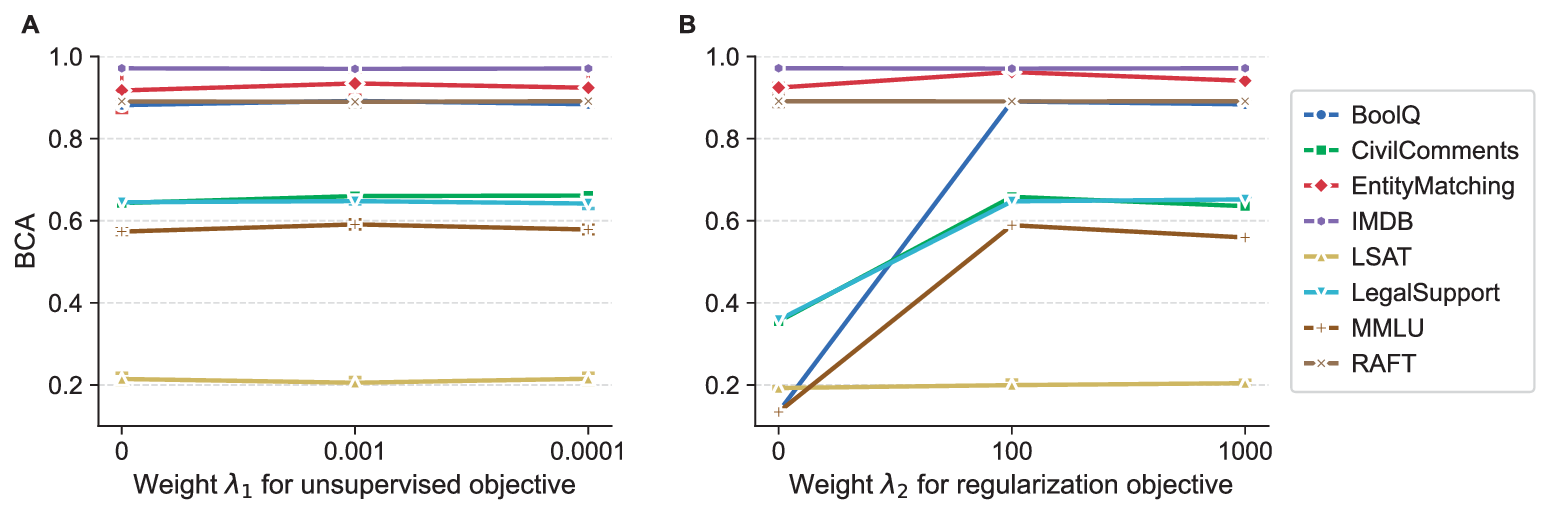}
\caption{{\bfseries Sensitivity of StackingNet to different weights for loss integration for classification combination.}
{\bfseries a}, BCA of StackingNet w.r.t. the weight $\lambda_1$ for unsupervised loss.
{\bfseries b}, BCA of StackingNet w.r.t. the weight $\lambda_2$ for regularization loss.
The weight for the supervised objective is fixed at 1.}
\label{fig:weight-sensitivity}
\end{figure}

\clearpage

\begin{table}[htpb]
\caption{Predicting the peer consensus on paper rating.}
\label{tab:humanloo}
\begin{threeparttable}
\resizebox{0.8\textwidth}{!}{
\begin{tabular}{l|cc|ccc}
\toprule
& \multicolumn{2}{c|}{Single human reviewer} & \multicolumn{3}{c}{AI predictors} \\
Dataset & Self-inclusive & Held-out & Best single LLM & Averaging & StackingNet \\
\midrule
ICLR2025 & 0.946 & 1.268 & 1.248 & 1.846 & $\mathbf{1.123}_{\pm0.11}$ \\
ICLR2024 & 0.938 & 1.280 & 1.248 & 1.857 & $\mathbf{1.130}_{\pm0.14}$ \\
NeurIPS2024 & 0.780 & 1.041 & 1.092 & 1.430 & $\mathbf{0.789}_{\pm0.14}$ \\
NeurIPS2023 & 0.782 & 1.016 & 1.098 & 1.443 & $\mathbf{0.721}_{\pm0.10}$ \\
\bottomrule
\end{tabular}}
\begin{tablenotes}[flushleft]
\footnotesize
\item Entries are the mean absolute error against the consensus target, where lower is better.
\item Self-inclusive: a reviewer scored against a consensus that still includes their own rating.
\item Held-out: a reviewer scored against the mean rating of the paper's other reviewers only.
\item The best single LLM, averaging, and StackingNet are all scored against the held-out consensus.
\item StackingNet is trained few-shot on the held-out reviewer's rating, which is excluded from its scoring target.
\item Entries are bootstrap means over 50 resamples, and bold marks the lowest held-out error.
\end{tablenotes}
\end{threeparttable}
\end{table}

\clearpage

\begin{table}[htpb] \centering \renewcommand{\arraystretch}{1.15} \setlength{\tabcolsep}{2.2mm}
\caption{Inter-model error dependence per task.}
\label{tab:dependence}
\begin{threeparttable}
\resizebox{0.66\textwidth}{!}{\begin{tabular}{l|c|c|c}
\toprule
Task & Base models & Mean error correlation & Joint-error rate \\
\midrule
\multicolumn{4}{l}{\textit{Regression}} \\
CFD (13 attributes) & 10 & 0.39 & -- \\
ICLR2025 & 6 & 0.51 & -- \\
ICLR2024 & 6 & 0.52 & -- \\
NeurIPS2024 & 6 & 0.37 & -- \\
NeurIPS2023 & 6 & 0.36 & -- \\
\midrule
\multicolumn{4}{l}{\textit{Classification}} \\
BoolQ & 10 & 0.35 & 0.08 \\
CivilComments & 10 & 0.23 & 0.28 \\
EntityMatching & 10 & 0.14 & 0.08 \\
IMDB & 10 & 0.34 & 0.02 \\
LegalSupport & 10 & 0.25 & 0.23 \\
LSAT & 10 & 0.13 & 0.64 \\
MMLU & 10 & 0.41 & 0.36 \\
RAFT & 10 & 0.35 & 0.07 \\
\bottomrule
\end{tabular}}
\begin{tablenotes}[flushleft]
\footnotesize
\item Mean error correlation: the mean off-diagonal correlation between base-model errors, on the residual (prediction minus label) for regression and the error indicator (one if wrong, zero if right) for classification. The CFD value is averaged over its thirteen attributes.
\item Joint-error rate: the fraction of items that two base models get wrong together, averaged over all model pairs (the double-fault diversity measure of ref~\cite{Kuncheva2003}). It is defined only for the classification tasks, so the regression rows are marked ``--''.
\end{tablenotes}
\end{threeparttable}
\end{table}

\clearpage

\begin{table}[htpb]  \centering \setlength{\tabcolsep}{1.0mm}
\caption{Mean absolute error on the CFD dataset for ablation studies.}  \label{tab:cfdablation}
\resizebox{\textwidth}{!}{\begin{tabular}{l|cc|cc|cccc}
\toprule
\multirow{2}{*}{Dataset} & \multicolumn{2}{c|}{Linear Regression} & \multicolumn{2}{c|}{StackingNet w/o bias} & \multicolumn{4}{c}{StackingNet w/ bias} \\ \cmidrule{2-3} \cmidrule{4-5} \cmidrule{6-9}
& w/o clamp & weight clamp & w/o clamp & weight clamp & w/o clamp & weight clamp & bias clamp & both clamp \\
\midrule
CFD-afraid & 1.399 & 0.512 & 1.406 & 0.512 & 1.406 & 0.513 & 1.404 & \textbf{0.510} \\
CFD-angry & 1.031 & 0.538 & 0.974 & 0.556 & 0.941 & 0.526 & 0.941 & \textbf{0.524} \\
CFD-attractive & 0.755 & 0.609 & 0.764 & 0.608 & 0.760 & 0.608 & 0.759 & \textbf{0.606} \\
CFD-babyfaced & 1.032 & 0.484 & 0.722 & 0.484 & 0.730 & 0.484 & 0.728 & \textbf{0.482} \\
CFD-disgusted & 2.311 & 0.611 & 2.115 & 0.543 & 2.137 & 0.611 & 2.123 & \textbf{0.508} \\
CFD-feminine & 0.743 & 0.509 & 0.736 & \textbf{0.505} & 0.734 & \textbf{0.505} & 0.736 & 0.507 \\
CFD-happy & 1.202 & \textbf{0.526} & 0.822 & 0.606 & 0.697 & \textbf{0.526} & 0.695 & 0.534 \\
CFD-masculine & 0.634 & 0.525 & 0.631 & 0.416 & 0.632 & 0.520 & 0.631 & \textbf{0.408} \\
CFD-sad & 0.832 & \textbf{0.514} & 0.837 & 0.570 & 0.830 & 0.549 & 0.830 & 0.548 \\
CFD-surprised & 1.235 & 1.114 & 1.056 & 0.591 & 1.026 & 1.113 & 1.016 & \textbf{0.579} \\
CFD-threatening & 5.743 & 0.751 & 1.444 & 0.540 & 2.579 & 0.751 & 1.443 & \textbf{0.537} \\
CFD-trustworthy & 3.396 & \textbf{0.395} & 0.792 & 0.587 & 0.576 & \textbf{0.395} & 0.571 & 0.472 \\
CFD-unusual & 0.942 & 0.772 & 0.855 & 0.517 & 0.876 & 0.768 & 0.852 & \textbf{0.455} \\
\midrule
Avg. & 1.635 & 0.605 & 1.012 & 0.541 & 1.071 & 0.979 & 0.605 & \textbf{0.513} \\
\bottomrule
\end{tabular}}
\begin{tablenotes}
\footnotesize
\item \qquad \textit{Clamp refers to non-negativity constraint.}
\end{tablenotes}
\end{table}

\clearpage

\begin{table}[htpb]  \centering
\caption{Complete information of the queried LLMs/VLMs.}  \label{tab:llmstatistics}
\begin{threeparttable}
\resizebox{\textwidth}{!}{\begin{tabular}{llllll}
\toprule
Model Alias & Complete Version ID & Organization & Parameters & Access / Query Date & URL \\
\midrule
DeepSeek-R1 & DeepSeek-R1-0528 & DeepSeek & 671 B & Jul. 17, 2025 & \href{https://api-docs.deepseek.com/}{DeepSeek} \\
Doubao & doubao-seed-1-6-250615 & ByteDance & Not specified & Jul. 7, 2025 & \href{https://www.volcengine.com/docs/82379/1494384}{ByteDance} \\
Gemini-2 & gemini-2.0-flash-001 & Google & Not specified & Jul. 15, 2025 & \href{https://ai.google.dev/gemini-api/docs/text-generation}{GoogleAPIs} \\
GPT-4o & gpt-4o-2024-08-06 & OpenAI & Not specified & Jul. 13, 2025 & \href{https://platform.openai.com/docs/overview}{OpenAI} \\
GPT-5 & gpt-5-2025-08-07 & OpenAI & Not specified & Aug. 11, 2025 & \href{https://platform.openai.com/docs/overview}{OpenAI} \\
Qwen-Turbo & qwen-turbo-2025-04-28 & Alibaba Cloud & Not specified & Jul. 18, 2025 & \href{https://cn.aliyun.com/product/tongyi?from_alibabacloud=}{AliYun} \\
\midrule
BLIP & blip-vqa-base & Salesforce & 385 M & Jun. 3, 2025 & \href{https://huggingface.co/Salesforce/blip-vqa-base}{Hugging Face} \\
DeepSeek-VL & deepseek-vl-7b-chat & DeepSeek & 7 B & Jun. 3, 2025 & \href{https://huggingface.co/deepseek-ai/deepseek-vl-7b-chat}{Hugging Face} \\
H2OVL & h2ovl-mississippi-2b & H2O.ai & 2 B & Jun. 3, 2025 & \href{https://huggingface.co/h2oai/h2ovl-mississippi-2b}{Hugging Face} \\
InternVL-2 & InternVL2-8B & OpenGVLab & 8 B & Jun. 3, 2025 & \href{https://huggingface.co/OpenGVLab/InternVL2-8B}{Hugging Face} \\
LLaVA & llava-onevision-qwen2-7b-si-hf & NTU \& ByteDance & 7 B & Jun. 3, 2025 & \href{https://huggingface.co/lmms-lab/llava-onevision-qwen2-7b-ov}{Hugging Face} \\
Molmo & Molmo-7B-O-0924 & AI2 & 7 B & Jun. 3, 2025 & \href{https://huggingface.co/allenai/Molmo-7B-O-0924}{Hugging Face} \\
Paligemma & paligemma-3b-mix-448 & Google & 3 B & Jun. 3, 2025 & \href{https://huggingface.co/google/paligemma-3b-mix-448}{Hugging Face} \\
Phi-3.5 & Phi-3.5-vision-instruct & Microsoft & 4 B & Jun. 3, 2025 & \href{https://huggingface.co/microsoft/Phi-3.5-vision-instruct}{Hugging Face} \\
SAIL-VL & SAIL-VL-1d5-2B & ByteDance & 2 B & Jun. 3, 2025 & \href{https://huggingface.co/BytedanceDouyinContent/SAIL-VL-1d5-2B}{Hugging Face} \\
SmolVLM & SmolVLM-Instruct & Hugging Face & 2 B & Jun. 3, 2025 & \href{https://huggingface.co/HuggingFaceTB/SmolVLM-Instruct}{Hugging Face} \\
\midrule
AnthropicLM-v4 & anthropic\_stanford-online-all-v4-s3 & Anthropic \& Stanford & 52 B & Nov. 17, 2023 & \href{https://www.anthropic.com/news/claude-3-family}{Anthropic} \\
Command & cohere\_command-xlarge-beta & Cohere & 52.4 B & Nov. 17, 2023 & \href{https://docs.cohere.com/docs/command-beta}{Cohere} \\
Falcon & tiiuae\_falcon-40b & Technology Innovation Institute & 40 B & Nov. 17, 2023 & \href{https://huggingface.co/tiiuae/falcon-40b}{Hugging Face} \\
GPT-3.5-Turbo & openai\_gpt-3.5-turbo-0301 & OpenAI & Not specified & Nov. 17, 2023 & \href{https://platform.openai.com/docs/overview}{OpenAI} \\
Jurassic-2 & ai21\_j2-jumbo & AI21 Labs & 178 B & Nov. 17, 2023 & \href{https://www.ai21.com/blog/introducing-j2/}{AI21} \\
Llama-2 & meta\_llama-2-70b & Meta & 70 B & Nov. 17, 2023 & \href{https://huggingface.co/meta-llama/Llama-2-70b-hf}{Hugging Face} \\
MPT & mosaicml\_mpt-instruct-30b & MosaicML & 30 B & Nov. 17, 2023 & \href{https://huggingface.co/mosaicml/mpt-30b-instruct}{Hugging Face} \\
Palmyra-X & writer\_palmyra-x & Writer & 43 B & Nov. 17, 2023 & \href{https://writer.com/llms/}{Writer} \\
RedPajama & together\_redpajama-incite-instruct-7b & Together AI & 7 B & Nov. 17, 2023 & \href{https://huggingface.co/togethercomputer/RedPajama-INCITE-7B-Instruct}{Hugging Face} \\
TNLG-v2 & microsoft\_TNLGv2\_530B & Microsoft \& Nvidia & 530 B & Nov. 17, 2023 & \href{https://www.microsoft.com/en-us/research/blog/turing-nlg-a-17-billion-parameter-language-model-by-microsoft/}{Microsoft} \\
\bottomrule
\end{tabular}}
\begin{tablenotes}
\footnotesize
\item \textit{LLMs/VLMs are ordered alphabetically.\\All queries for the datasets from HELM benchmark were derived from their official release.}
\end{tablenotes}
\end{threeparttable}
\end{table}

\clearpage

\textbf{Prompts for Research Paper Rating} The prompt follows a structured format comprising several components: it assigns the model the role of a reviewer, provides the official review guidelines of the target conference, includes the full manuscript text, and requests both a detailed qualitative assessment and quantitative ratings. The design builds on the structured prompting framework introduced by Yu~\emph{et al.}~\cite{Yu2025}. For the regression task, only the scalar rating is used as supervision. By incorporating these contextual elements, the prompt encourages more grounded and consistent evaluations of complex academic manuscripts, aligning the model's reasoning more closely with the cognitive process of human reviewers. The complete prompt format is shown below.

\begin{tcolorbox}[colframe=blue!80!black, colback=blue!10!white, title=LLM prompt for research paper rating, breakable]

\textit{[System prompt]}

\textit{You are an AI researcher reviewing a paper submitted to a prestigious AI research conference. You will be provided with the manuscript text, the conference's reviewer guidelines and templates. Your objective is to thoroughly evaluate the paper, adhering to the provided guidelines and the specified response template. Ensure your evaluation is objective, comprehensive, and aligned with the conference standards.}

\textit{\#\# Reviewer Guidelines}

\textit{\{Reviewer Guidelines of the Conference\}}

\textit{\#\# Response Template (JSON format)}

\textit{Provide the review in valid JSON format with the following fields. Ensure all fields are completed as described below. The response must be a valid JSON object.}

\textit{- "summary\_of\_the\_paper": Briefly summarize the paper and its contributions. This is not the place to critique the paper; the authors should generally agree with a well-written summary. You may use paragraphs and bulleted lists for formatting, but ensure that the content remains a single, continuous text block. Do not use nested JSON or include additional fields.}

\textit{- "main\_review": Provide review comments as a single text field (a string). Consider including assessment on the following dimensions: a comprehensive list of strong and weak points of the paper, your recommendation, supporting arguments for your recommendation, questions to clarify your understanding of the paper or request additional evidence, and additional feedback with the aim to improve the paper. You may use paragraphs and bulleted lists for formatting, but ensure that the content remains a single, continuous text block. Do not use nested JSON or include additional fields.}

\textit{- "summary\_of\_the\_review": Concise summary of `main\_review'. You may use paragraphs and bulleted lists for formatting, but ensure that the content remains a single, continuous text block. Do not use nested JSON or include additional fields.}

\textit{- "correctness": A numerical rating on the following scale to indicate that the claims and methods are correct. The value should be between 1 and 4, where:}
\begin{itemize}
\item \textit{1 = The main claims of the paper are incorrect or not at all supported by theory or empirical results.}
\item \textit{2 = Several of the paper's claims are incorrect or not well-supported.}
\item \textit{3 = Some of the paper's claims have minor issues. A few statements are not well-supported, or require small changes to be made correct.}
\item \textit{4 = All of the claims and statements are well-supported and correct.}
\end{itemize}

\textit{- "technical\_novelty\_and\_significance": A numerical rating on the following scale to indicate technical novelty and significance. The value should be between 1 and 4, where:}
\begin{itemize}
\item \textit{1 = The contributions are neither significant nor novel.}
\item \textit{2 = The contributions are only marginally significant or novel.}
\item \textit{3 = The contributions are significant and somewhat new. Aspects of the contributions exist in prior work.}
\item \textit{4 = The contributions are significant and do not exist in prior works.}
\end{itemize}

\textit{- "empirical\_novelty\_and\_significance": A numerical rating on the following scale to indicate empirical novelty and significance. The value should be between 1 and 4, or -999 if not applicable, where:}
\begin{itemize}
\item \textit{1 = The contributions are neither significant nor novel.}
\item \textit{2 = The contributions are only marginally significant or novel.}
\item \textit{3 = The contributions are significant and somewhat new. Aspects of the contributions exist in prior work.}
\item \textit{4 = The contributions are significant and do not exist in prior works.}
\item \textit{-999 = Not applicable.}
\end{itemize}

\textit{- "flag\_for\_ethics\_review": A boolean value (`true' or `false') indicating whether there are ethical concerns in the work.}

\textit{- "recommendation": A string indicating the final decision, which must strictly be one of the following options: `strong reject', `reject, not good enough', `marginally below the acceptance threshold', `marginally above the acceptance threshold', `accept, good paper', or `strong accept, should be highlighted at the conference'.}

\textit{- "rating": A float indicating the final decision score in the range [1, 10] that corresponds to the recommendation, where higher scores indicate better paper.}

\textit{- "confidence": A numerical values to indicate how confident you are in your evaluation. The value should be between 1 and 5, where:}
\begin{itemize}
\item \textit{1 = You are unable to assess this paper and have alerted the ACs to seek an opinion from different reviewers.}
\item \textit{2 = You are willing to defend your assessment, but it is quite likely that you did not understand the central parts of the submission or that you are unfamiliar with some pieces of related work. Math/other details were not carefully checked.}
\item \textit{3 = You are fairly confident in your assessment. It is possible that you did not understand some parts of the submission or that you are unfamiliar with some pieces of related work. Math/other details were not carefully checked.}
\item \textit{4 = You are confident in your assessment, but not absolutely certain. It is unlikely, but not impossible, that you did not understand some parts of the submission or that you are unfamiliar with some pieces of related work.}
\item \textit{5 = You are absolutely certain about your assessment. You are very familiar with the related work and checked the math/other details carefully.}
\end{itemize}

\textit{[User prompt]}

\textit{Here is the paper you are asked to review. Write a well-justified review of this paper:}

\textit{\{Full paper text converted from the PDF source\}}

\end{tcolorbox}

We used the official OpenReview API (\url{https://github.com/openreview/openreview-py}) to retrieve paper PDFs and corresponding human reviewer overall rating scores. To convert to text, the documents were then parsed using the PyMuPDF (\url{https://github.com/pymupdf/PyMuPDF}) Python library to extract markdown-style text. To simulate an anonymous peer review setting, we applied regular expression filters to systematically remove identifying details, including author names, institutional affiliations, contact information, funding declarations, publication metadata, etc. The scalar rating values returned by the models were used as regression scores.

\textbf{Prompts for Facial Image Attribute Rating} The prompt is designed to elicit judgments of perceived facial attributes without introducing contextual information, assigned roles, or preference framing. It presents a direct, neutrally worded brief question following the image token input.

\begin{tcolorbox}[colframe=blue!80!black, colback=blue!10!white, title=VLM prompt for facial image attribute rating, breakable]

\textit{\{Image content represented as converted tokens\}}

\textit{On a scale of 1 to 7, how \{Adjective\} does the person pictured above appear? (1 = Not at all, 4 = Neutral, 7 = Extremely) }

\textit{Answer in a single float:}

\end{tcolorbox}

\textbf{Prompts for HELM Datasets}
Each of the eight datasets in the HELM benchmark provides a predefined textual query. Representative examples are listed in Supporting Information Table~\ref{tab:helmdatasets}.

\begin{table}[htpb] \centering
\caption{Example query prompts from datasets of the HELM benchmark.}  \label{tab:helmdatasets}
\begin{threeparttable}
\resizebox{\textwidth}{!}{\begin{tabular}{ll}
\toprule
Dataset Name & Example \\
\midrule
\multirow{5}{*}{BoolQ} & \multirow{5}{*}{\shortstack[l]{\textit{Passage: The center contact of the bulb typically connects to the medium-power filament, and the ring connects to the low-power filament. }\\\textit{Thus, if a 3-way bulb is screwed into a standard light socket that has only a center contact, only the medium-power filament operates.} \\\textit{In the case of the 50 W / 100 W / 150 W bulb, putting this bulb in a regular lamp socket will result in it behaving like a normal 100W bulb.} \\ \textit{Question: Do 3 way light bulbs work in any lamp?} \\ \textit{Answer:}}} \\
\\
\\
\\
\\
\midrule
\multirow{3}{*}{CivilComments} & \multirow{3}{*}{\shortstack[l]{\textit{Passage: she'll be whatever she needs to be in order to gain office. once she's there she'll not give a crap what you think.} \\ \textit{Question: is the passage above toxic?} \\ \textit{Answer:}}} \\
\\
\\
\midrule
\multirow{5}{*}{EntityMatching} & \multirow{5}{*}{\shortstack[l]{\textit{Are Product A and Product B the same? Yes or No?} \\ \textit{Product A is Beer\_Name: Fisherman 's Tea Party. Brew\_Factory\_Name: Cape Ann Brewing Company. Style: American Barleywine. ABV: 9.00 \%.} \\\textit{Product B is Beer\_Name: Cape Ann Fisherman\_s Tea Party. Brew\_Factory\_Name: Cape Ann Brewing Company. Style: Barley Wine. ABV: 9 \%.} \\ \textit{Are A and B the same?} \\ \textit{Answer:}}} \\
\\
\\
\\
\\
\midrule
\multirow{4}{*}{IMDB} & \multirow{4}{*}{\shortstack[l]{\textit{Passage: This film is totally mindblowing. It manages to be thought provoking, funny, tragic, and cinematic yet claustrophobic.} \\\textit{Although the flashbacks are unnecessary, the film maintains a pacy, punchy grip and the performances are all excellent,} \\ \textit{in particular Alec Baldwin, and the mesmerising Eric Bogosian as the film's anti-hero, Barry Champlain.} \\ \textit{Sentiment:}}} \\
\\
\\
\\
\midrule
\multirow{5}{*}{LegalSupport} & \multirow{5}{*}{\shortstack[l]{\textit{Which statement best supports the passage?} \\ \textit{Passage: A truly vindictive prosecution is illegal because it violates due process.} \\\textit{A. federal prosecution may proceed if federal prosecutor did not participate in state prosecutor's allegedly vindictive action against defendant.} \\\textit{B. vindictiveness may play no part in resentencing of criminal defendant who has successfully invoked right to appeal.} \\\textit{Answer:}}} \\
\\
\\
\\
\\
\midrule
\multirow{6}{*}{LSAT} & \multirow{6}{*}{\shortstack[l]{\textit{The following are multiple choice questions. (with answers)} \\ \textit{Bird-watchers explore a forest to see which of the following six kinds of birds-grosbeak, harrier, jay, martin, shrike, wren-it contains. The findings} \\ \textit{are consistent with the following conditions: If harriers are in the forest, then grosbeaks are not. If jays, martins, or both are in the forest, then so} \\ \textit{are harriers. If wrens are in the forest, then so are grosbeaks. If jays are not in the forest, then shrikes are.} \\ \textit{Question: Which one of the following is the maximum number of the six kinds of birds the forest could contain? A. two B. three C. four D. five E. six}\\ \textit{Answer:}}} \\
\\
\\
\\
\\
\\
\midrule
\multirow{4}{*}{MMLU} & \multirow{4}{*}{\shortstack[l]{\textit{The following are multiple choice questions (with answers) about computer security.}\\ \textit{Question: The \_\_\_\_\_\_\_\_\_\_\_\_ is anything which your search engine cannot search.} \\ \textit{A. Haunted web B. World Wide Web C. Surface web D. Deep Web}\\ \textit{Answer:}}} \\
\\
\\
\\
\midrule
\multirow{7}{*}{RAFT} & \multirow{7}{*}{\shortstack[l]{\textit{Label whether the following tweet contains hate speech against either immigrants or women. Hate Speech (HS) is commonly defined as any} \\ \textit{communication that disparages a person or a group on the basis of some characteristic such as race, color, ethnicity, gender, sexual orientation,} \\ \textit{nationality, religion, or other characteristics.} \\ \textit{Possible labels: 1. hate speech 2. not hate speech} \\ T\textit{weet: It's clear dat the administration's attempts to punish local governments fa refusing to cooperate wit immigration enforcement r} \\ \textit{unconstitutional. Da administration should focus on treating immigrants w/ compassion and respect.} \\ \textit{Label:}}} \\
\\
\\
\\
\\
\\
\\
\bottomrule
\end{tabular}}
\begin{tablenotes}
\footnotesize
\item \textit{Each full query includes few-shot exemplars with answers.\\For clarity and brevity, only the final query prompt of a single sample is shown.}
\end{tablenotes}
\end{threeparttable}
\end{table}

\end{document}